\documentclass{article}

\usepackage[english]{babel}

\usepackage[letterpaper,top=2cm,bottom=2cm,left=3cm,right=3cm,marginparwidth=1.75cm]{geometry}

\usepackage{graphicx}
\usepackage[colorlinks=true, allcolors=blue]{hyperref}
\usepackage{amsmath,amsfonts,amsthm}
\usepackage{algorithmic}
\usepackage{algorithm}

\usepackage{textcomp}
\usepackage{stfloats}
\usepackage{url}
\usepackage{verbatim}
\usepackage{cite}
\usepackage{comment}

\def\cast{{
   \mathord{
      \hbox to 0em{
         \ooalign{
	   \smash{\hbox{$\ast$}}\crcr
	   \smash{\hskip-1pt\Large\hbox{$\circ$}} }
	 \hidewidth}
      \phantom{\bigcirc}
} }}

\newcommand{\rT}{^{ \raisebox{1.2pt}{$\rm \scriptstyle T$}}}

\newcommand{\bds}{\begin {itemize}}
\newcommand{\eds}{\end {itemize}}
\newcommand{\bdf}{\begin{definition}}
\newcommand{\blm}{\begin{lemma}}
\newcommand{\edf}{\end{definition}}
\newcommand{\elm}{\end{lemma}}
\newcommand{\bthm}{\begin{theorem}}
\newcommand{\ethm}{\end{theorem}}
\newcommand{\bprp}{\begin{prop}}
\newcommand{\eprp}{\end{prop}}
\newcommand{\bcl}{\begin{claim}}
\newcommand{\ecl}{\end{claim}}
\newcommand{\bcr}{\begin{coro}}
\newcommand{\ecr}{\end{coro}}
\newcommand{\bquest}{\begin{question}}
\newcommand{\equest}{\end{question}}

\newcommand{\larrow}{{\larrow}}

\newcommand{\argmin}{\ensuremath{\mathrm{arg}\min}}
\newcommand{\argmax}{\ensuremath{\mathrm{arg}\max}}

\newcommand{\vb}{{\ensuremath{{\mathbf{b}}}}}

\newcommand{\ve}{{\ensuremath{{\mathbf{e}}}}}

\newcommand{\vh}{{\ensuremath{{\mathbf{h}}}}}

\newcommand{\vs}{{\ensuremath{{\mathbf{s}}}}}

\newcommand{\vw}{{\ensuremath{{\mathbf{w}}}}}

\newcommand{\vx}{{\ensuremath{{\mathbf{x}}}}}

\newcommand{\vy}{{\ensuremath{{\mathbf{y}}}}}
\newcommand{\vz}{{\ensuremath{{\mathbf{z}}}}}

\newcommand{\mA}{{\ensuremath{\mathbf{A}}}}

\newcommand{\mB}{{\ensuremath{\mathbf{B}}}}

\newcommand{\mD}{{\ensuremath{\mathbf{D}}}}

\newcommand{\mI}{{\ensuremath{\mathbf{I}}}}

\newcommand{\mW}{{\ensuremath{\mathbf{W}}}}

\newcommand{\mX}{{\ensuremath{\mathbf{X}}}}

\newcommand{\mZ}{{\ensuremath{\mathbf{Z}}}}

\usepackage{latexsym}

\def\IC{\mathbb C}
\def\IN{\mathbb N}
\def\IZ{\mathbb Z}
\def\IR{\mathbb R}

\def\shat{^{\mathchoice{}{}%
 {\,\,\smash{\hbox{\lower4pt\hbox{$\widehat{\null}$}}}}%
 {\,\smash{\hbox{\lower3pt\hbox{$\hat{\null}$}}}}}}

\def\bSigma{{
      \ooalign{
      \smash{\hskip.4pt\raise.4pt\hbox{$\Sigma$}}\vphantom{}\crcr
      \smash{\hskip.7pt\raise.6pt\hbox{$\Sigma$}}\vphantom{}\crcr
      \smash{\hbox{$\Sigma$}}\vphantom{$\Sigma$}}
      \vphantom{\hbox{$\Sigma$}}
      }}
\def\bTheta{{
      \ooalign{
      \smash{\hskip.5pt\raise.5pt\hbox{$\Theta$}}\vphantom{}\crcr
      \smash{\hskip.0pt\raise.1pt\hbox{$\Theta$}}\vphantom{}\crcr
      \smash{\hbox{$\Theta$}}\vphantom{$\Theta$}}
      \vphantom{\hbox{$\Theta$}}
      }}
\def\bDelta{{
      \ooalign{
      \smash{\hskip.4pt\raise.4pt\hbox{$\Delta$}}\vphantom{}\crcr
      \smash{\hskip.7pt\raise.6pt\hbox{$\Delta$}}\vphantom{}\crcr
      \smash{\hbox{$\Delta$}}\vphantom{$\Delta$}}
      \vphantom{\hbox{$\Delta$}}
      }}
\def\bLambda{{
      \ooalign{
      \smash{\hskip.5pt\raise.5pt\hbox{$\Lambda$}}\vphantom{}\crcr
      \smash{\hskip.0pt\raise.1pt\hbox{$\Lambda$}}\vphantom{}\crcr
      \smash{\hbox{$\Lambda$}}\vphantom{$\Lambda$}}
      \vphantom{\hbox{$\Lambda$}}
      }}

\makeatletter

\def\bordermatrix#1{\begingroup \m@th
  \@tempdima 8.75\p@
  \setbox\z@\vbox{%
    \def\cr{\crcr\noalign{\kern2\p@\global\let\cr\endline}}%
    \ialign{$##$\hfil\kern2\p@\kern\@tempdima&\thinspace\hfil$##$\hfil
      &&\quad\hfil$##$\hfil\crcr
      \omit\strut\hfil\crcr\noalign{\kern-\baselineskip}%
      #1\crcr\omit\strut\cr}}%
  \setbox\tw@\vbox{\unvcopy\z@\global\setbox\@ne\lastbox}%
  \setbox\tw@\hbox{\unhbox\@ne\unskip\global\setbox\@ne\lastbox}%
  \setbox\tw@\hbox{$\kern\wd\@ne\kern-\@tempdima\left[\kern-\wd\@ne
    \global\setbox\@ne\vbox{\box\@ne\kern2\p@}%
    \vcenter{\kern-\ht\@ne\unvbox\z@\kern-\baselineskip}\,\right]$}%
  \null\;\vbox{\kern\ht\@ne\box\tw@}\endgroup}
\makeatother

\makeatletter
\def\argmin{\mathop{\operator@font arg\,min}}
\def\argmax{\mathop{\operator@font arg\,max}}
\makeatother

\newcommand{\bea}{\begin{array}}
\newcommand{\ena}{\end{array}}
\newcommand{\beq}{\begin{equation}}
\newcommand{\enq}{\end{equation}}

\newcommand{\beqa}{\begin{eqnarray}}
\newcommand{\enqa}{\end{eqnarray}}

\newcommand{\beqan}{\begin{eqnarray*}}
\newcommand{\enqan}{\end{eqnarray*}}

\newcommand{\AL}{\begin{enumerate}}
\newcommand{\ALE}{\end{enumerate}}

\def\addots{\mathinner{
    \mkern1mu\raise0pt\vbox{\kern7pt\hbox{.}}
    \mkern2mu\raise4pt\hbox{.}
    \mkern2mu\raise7pt\hbox{.}
    \mkern1mu}}

\def\sddots{\mathinner{
    \mkern.8mu\raise7pt\hbox{.}
    \mkern.8mu\raise4pt\hbox{.}
    \mkern.8mu\raise0pt\vbox{\kern7pt\hbox{.}}
    \mkern1mu}}

\def\saddots{\mathinner{
    \mkern.2mu\raise0pt\vbox{\kern7pt\hbox{.}}
    \mkern.2mu\raise4pt\hbox{.}
    \mkern.2mu\raise7pt\hbox{.}
    \mkern1mu}}

\def\sqplus{\mathbin{
	{\ooalign{\hfil\raise.3ex\hbox{\scriptsize
	+}\hfil\crcr\mathhexbox274\crcr\mathhexbox275}}
	}} 
\def\sqminus{\mathbin{
	{\ooalign{\hfil\raise.3ex\hbox{\scriptsize
	--}\hfil\crcr\mathhexbox274\crcr\mathhexbox275}}
	}}

\def\IC{{
   \mathord{
      \hbox to 0em{
	 \hskip-4pt
         \ooalign{
	   \smash{\hskip1.9pt\raise2.6pt\hbox{$\scriptscriptstyle |$}}\crcr
	   \smash{\hbox{\rm\sf C}} }
	 \hidewidth}
      \phantom{\hbox{\rm\sf C}}
} }}
\def\IN{
    {\ooalign{
   \smash{\hskip2.2pt\raise1.5pt\hbox{$\scriptscriptstyle |$}}\vphantom{}\crcr
   \hbox{\sf N}
	}}
	} 
\def\IZ{
    {\ooalign{
   \smash{\hskip1.9pt\raise0pt\hbox{$\sf Z$}}\vphantom{}\crcr
   \hbox{\sf Z}
	}}
	} 
\def\IR{
    {\ooalign{
   \smash{\hskip2.2pt\raise1.5pt\hbox{$\scriptscriptstyle |$}}\vphantom{}\crcr
   \smash{\hskip2.2pt\raise3.3pt\hbox{$\scriptscriptstyle |$}}\vphantom{}\crcr
   \hbox{\sf R}
	}}
	} 

\DeclareMathAlphabet{\mathcmb}{OT1}{cmr}{b}{n}

\def\bSigma{\ensuremath{\mathcmb{\Sigma}}}
\def\bLambda{\ensuremath{\mathcmb{\Lambda}}}

\def\bTheta{\ensuremath{\mathcmb{\Theta}}}

\newcommand{\SI}{\begin{indlist}}
\newcommand{\EI}{\end{indlist}}

\newcommand{\DL}{\begin{dashlist}}
\newcommand{\DLE}{\end{dashlist}}

\makeatletter
\def\setboxz@h{\setbox\z@\hbox}
\def\wdz@{\wd\z@}
\def\boxz@{\box\z@}
\def\underset#1#2{\binrel@{#2}%
  \binrel@@{\mathop{\kern\z@#2}\limits_{#1}}}
\def\binrel@#1{\begingroup
  \setboxz@h{\thinmuskip0mu
    \medmuskip\m@ne mu\thickmuskip\@ne mu
    \setbox\tw@\hbox{$#1\m@th$}\kern-\wd\tw@
    ${}#1{}\m@th$}%
  \edef\@tempa{\endgroup\let\noexpand\binrel@@
    \ifdim\wdz@<\z@ \mathbin
    \else\ifdim\wdz@>\z@ \mathrel
    \else \relax\fi\fi}%
  \@tempa
}
\let\binrel@@\relax%
\makeatother

\newtheorem{theorem}{Theorem}
\newtheorem{lemma}{Lemma}

\newtheorem{prop}{Proposition}
\usepackage{subcaption}

\newtheorem*{definition}{Definition}

\usepackage{array}
\renewcommand{\arraystretch}{1.2}
\usepackage{hyperref}
\usepackage{xcolor}

\title{Graph Neural Networks with Parallel Neighborhood Aggregations for Graph Classification}
\author{Siddhant~Doshi
        and Sundeep~Prabhakar~Chepuri, \emph{IEEE Member}
        }

\usepackage{blindtext}
\newcommand\blfootnote[1]{%
\begingroup
\renewcommand\thefootnote{}\footnote{#1}%
\addtocounter{footnote}{-1}%
\endgroup
}

\begin{document}
\maketitle

\begin{abstract}
We focus on graph classification using a graph neural network (GNN) model that precomputes the node features using a bank of neighborhood aggregation graph operators arranged in parallel. These GNN models have a natural advantage of reduced training and inference time due to the precomputations but are also fundamentally different from popular GNN variants that update node features through a sequential neighborhood aggregation procedure during training. We provide theoretical conditions under which a generic GNN model with parallel neighborhood aggregations (PA-GNNs, in short) are provably as powerful as the well-known Weisfeiler-Lehman (WL) graph isomorphism test in discriminating non-isomorphic graphs.  Although PA-GNN models do not have an apparent relationship with the WL test, we show that the graph embeddings obtained from these two methods are injectively related. We then propose a specialized PA-GNN model, called SPIN, which obeys the developed conditions. We demonstrate via numerical experiments that the developed model achieves state-of-the-art performance on many diverse real-world datasets while maintaining the discriminative power of the WL test and the computational advantage of preprocessing graphs before the training process.
\end{abstract}

\section{Introduction}\label{sec:intro}
Graph neural networks (GNNs) have recently emerged as one the most popular machine learning models for processing and analyzing graph-structured data~\cite{GDL,gamma20spm}. GNNs have gained significant and steady attention due to their extraordinary success in solving many challenging tasks in a variety of scientific disciplines such as computational pharmacology~\cite{decagon}, molecular chemistry~\cite{MoIGAN}, physics~\cite{phy}, finance~\cite{finance1,finance2}, wireless communications~\cite{wicom}, and combinatorial optimization~\cite{opt}, to name a few.

\blfootnote{The authors are with the Department
of Electrical Communication Engineering, Indian Institute of Science, Bangalore,  560012, India.\\ 
Email: siddhant.doshi@outlook.com, spchepuri@iisc.ac.in \\
This work was supported in part by the SERB SRG/2019/000619 grant and Pratiksha Trust Fellowship.}
Solving machine learning tasks with graphs such as node property prediction, link prediction, or graph property prediction requires an efficient representation of the underlying graph~\cite{dong_gsp,graphBook}. In this work, we focus on semi-supervised graph classification, wherein we are given multiple graphs, each with an associated categorical label and data about the nodes in each graph. The goal is to train a machine learning model that processes the graph-structured data and nodal attributes (i.e., signals associated with the nodes) to predict labels of unseen graphs. For example, multiple graphs may represent different communities of people in a social network (chemical compounds), and the task is to identify the type of community (respectively, enzyme or not). 

\subsection{Related works}
A majority of GNNs learn the representation vector of a node (or node embedding) through a sequential neighborhood aggregation procedure during the training process. In each iteration (also referred to as a GNN layer), the representation vector of a node is computed by aggregating representation vectors from its local one-hop neighbors via a first-order graph filtering operation~\cite{gamma20spm}. Cascading many such layers with non-linear activation functions allows GNNs to learn the structural information beyond the node's one-hop neighborhood and thus achieve good generalization. Let us call GNN models that sequentially aggregate the neighbor node embeddings as {\it SA-GNNs} (here, SA stands for sequential aggregation). Different choices of sequential neighborhood aggregation functions lead to popular GNN variants such as graph convolutional networks (GCN)~\cite{GCN}, GraphSAGE~\cite{GraphSAGE}, graph attention networks (GAT)~\cite{GAT}, and graph isomorphism networks (GIN)~\cite{GIN}. Such SA-GNN models can suffer from gradient issues, over smoothing, or the so-called bottleneck effects~\cite{GCN},~\cite{Li18deepGCN}. Further, for graph classification, the representation vectors of all the nodes in a graph have to be pooled (or readout) to obtain a representation vector of an entire graph, and is referred to as graph embedding. Different graph pooling schemes have been proposed for graph classification with SA-GNNs~\cite{SAG,diffpool,DGCNN,ECC}. 

Recently, GNN models that precompute nodal features by performing neighborhood aggregation at multiple scales in a non-sequential or parallel manner through integer powers of graph operators have been proposed. Examples of such architectures include scalable inception graph neural networks (SIGN)~\cite{SIGN} and graph-augmented MLPs (GA-MLPs)~\cite{GAMLP}. These GNN architectures are analogous to inception modules for convolutional neural networks (CNNs), where a filterbank with convolutional filters of different sizes is used~\cite{inception}. Let us call GNN models in which a bank of non-sequential or parallel neighborhood aggregation functions, i.e., \emph{graph filterbanks}~\cite{Tremblay17graphfilterbank}, that gather node features from different neighborhoods as {\it PA-GNNs} (here, PA stands for parallel aggregation). Since nodal features in PA-GNNs are precomputed, these models benefit from reduced training and inference time compared to SA-GNNs while preserving the structural information about the underlying graph. A natural question to ask is, how do PA-GNN models perform on machine learning tasks with graphs? 

While experimental results for node prediction tasks have been presented with SIGN~\cite{SIGN} and GA-MLP~\cite{GAMLP} architectures, we focus on graph classification tasks with PA-GNN models in this work. For graph classification tasks with GNNs, understanding the discriminative power of the model plays a crucial role. Although most of the neural network models are usually developed based on empirical and experimental evidence, there are theoretical results available to characterize and analyze the discriminative power of a few popular GNN variants, mostly SA-GNNs, to distinguish two different non-isomorphic graph structures~\cite{GIN,Morris} by relating them to the well-known Weisfeiler-Lehman (WL) graph isomorphism test~\cite{WL_1968,babai}. Specifically, by leveraging the similarity in the neighborhood aggregation iterations in SA-GNNs to the vertex refinement updates in the WL graph isomorphism test, one can determine if a GNN model is as powerful as the WL test or build powerful GNN models such as GIN~\cite{GIN}. For instance, \cite{GIN} provides examples of node-level aggregation functions, which when used fail to distinguish graphs that the WL test would distinguish and the discriminative power of integer powers of graph operators is studied in \cite{GAMLP}. Graph classification based on SA-GNNs but without any theoretical characterization about their discriminative power are also commonly used~\cite{DGCNN,diffpool}. 
Non-GNN models such as graph kernel-based approaches~\cite{kernel1,social} for graph classification often do not scale well as computing the kernel matrix for a large number of graphs becomes intractable.

\subsection{Main results and contributions}
Unlike SA-GNNs, PA-GNN models considered in this work have several parallel branches, with each branch focusing on different neighborhood depths and yielding representation vectors of all the nodes in the graph. Therefore, to combine the representations from different branches, we propose a new architecture called \emph{simple and parallel graph isomorphism network} (SPIN). In SPIN, we perform branch-level readouts to obtain the representation vectors of an entire graph at every branch. These yield graph embeddings associated to different neighborhood depths. Further, we pool these branch-level graph representation vectors through a graph-level readout to obtain a single representation vector of an entire graph. We focus on studying the discriminative power of an end-to-end PA-GNN model and provide conditions on branch-level readout and graph-level readout functions under which PA-GNNs are as powerful as the WL test. Our main contributions and results are summarized as follows.

\begin{itemize}
    \item We propose a generic PA-GNN model, which aggregates node features from different neighborhoods in a non-sequential manner through a bank of graph operators arranged in parallel, pools node-level representations from each branch through branch-level readouts, and finally pools graph embeddings from multiple branches through a graph-level readout.
    \item We provide theoretical conditions on the branch-level and graph-level readouts under which PA-GNN models are as powerful as the WL test. Unlike the SA-GNN models, the procedure to compute graph embeddings in PA-GNN models do not admit the same form as the updates in the WL test. Towards this end, we show how the embeddings generated from PA-GNNs can be injectively mapped to the updates in the WL test, thereby maintaining the discriminative power of the 1-WL test.
    As a result, we show that simple graph-level readout functions, like max or mean graph-level readouts fail to discriminate regular graphs with the same node type.  
    \item Based on these generic conditions, we propose SPIN. Specifically, we present two variants of SPIN with and without an attention mechanism at the branch-level readouts to capture the most relevant node features at each branch. Further, we theoretically show that introducing an attention mechanism as a node-level function does not reduce the discriminative power of the model.
\end{itemize}

We validate the developed model and theory through extensive numerical experiments on twelve benchmark datasets for graph classification. These twelve datasets include five datasets from the social domain, five datasets from the chemical domain, and two datasets related to brain networks. We demonstrate that PA-GNN models with precomputed node features perform on par with SA-GNNs models while maintaining the discriminative power of the WL test and the computational advantage of preprocessing graphs before the training process.

\subsection{Organization}

The rest of the paper is organized as follows. In Section~\ref{sec:prelim}, we provide the relevant background on SA-GNN models and the WL test. In Section~\ref{sec:arch}, we present a generic PA-GNN model for learning graph embeddings and also provide a theoretical characterization of the PA-GNN model, which we specialize as SPIN in Section~\ref{sec:SPIN}. In Section~\ref{sec:exp}, we present results on a variety of datasets. The paper finally concludes in Section~\ref{sec:conclusions}.

Software and datasets to reproduce the results in the paper are available at \href{https://github.com/siddhant-doshi/SPIN}{\texttt{\url{https://github.com/siddhant-doshi/SPIN}}.}

\section{Preliminaries}\label{sec:prelim}
In this section, we briefly introduce a generic version of GNNs that update node features via sequential neighborhood aggregation. We then describe the Weisfeiler-Lehman graph isomorphism test, which is introduced to characterize the discriminative power of GNNs.

\subsection{Notation}
Let $\mathcal{G} = (\mathcal{V},\mathcal{E})$ denote a graph with node set $\mathcal{V}$ and edge set~$\mathcal{E}$. Each node $v \in \mathcal{V}$ has a feature (or attribute) vector $\vx_v \in \mathbb{R}^d$. A graph with $N$ nodes has an adjacency matrix $\mA \in \mathbb{R}^{N\times N}$ and input feature matrix $\mX = [\vx_1,\cdots,\vx_N]\rT \in \mathbb{R}^{N \times d}$. We denote the inner product between two vectors $\vx$ and $\vy$ as~$\langle\vx,\vy\rangle$. Let $\mathcal{N}_v^{(k)}$ denote the  set of $k$-hop neighboring nodes of node~$v \in \mathcal{V}$. We frequently use an extension of a set called multiset, which is defined as follows.

\begin{definition}[Multiset] Multiset is a collection of elements, which may occur more than once. It is a 2-tuple $(\mathcal{X},m)$, where $\mathcal{X}$ is the underlying set and $m: \cal{X} \mapsto \mathbb{Z}_+$ is a function that gives the multiplicity of each element $x \in \cal{X}$ as $m(x)$.
\end{definition}

\subsection{GNNs with sequential neighborhood aggregation} 

Most GNNs follow a sequential architecture comprising of a cascade of several local neighborhood aggregation layers, where each layer computes the representation vector of a node by aggregating feature vectors from its 1-hop neighboring nodes. Cascading $K$ such local aggregation layers captures the structural information within the $K$-hop neighborhood of a node. The procedure can be viewed as iteratively updating the representation vector of a node as 
\begin{equation}
\label{eq:SA-GCN}
	\vx_{v}^{(k+1)} = \phi^{(k)}\left(\vx_v^{(k)},f^{(k)}\left(\left\{\vx_i^{(k)},\; \forall\; i \in \mathcal{N}_{v}^{(1)}\right\}\right)\right),
\end{equation}
where $\vx_v^{(k)}$ is the $d_k$-dimensional representation vector of node $v$ at the $k$-th layer with $\vx_v^{(0)}$ being its input feature vector, $f^{(k)}(\cdot)$ is a graph operator that acts as the local aggregation function and propagates node features, and $\phi^{(k)}(\cdot)$ combines the neighborhood information of a node with its own representation vector. Several variants of SA-GNNs have been proposed with different choices of $f^{(k)}(\cdot)$ and $\phi^{(k)}(\cdot)$, such as GCN~\cite{GCN}, GraphSAGE~\cite{GraphSAGE}, and GIN~\cite{GIN}, to name a few. An example of a basic SA-GNN model in~\eqref{eq:SA-GCN} is
\[
\mX^{(k+1)} = \sigma\left(\mX^{(k)}\mW^{(k)}_{\rm self} + \mA\mX^{(k)}\mW^{(k)}_{\rm neigh}\right),
\]
where $\mX^{(k)} = [\vx_1^{(k)},\cdots,\vx_N^{(k)}]\rT \in \mathbb{R}^{N \times d_k}$. The matrices $\mW^{(k)}_{\rm self} \in \mathbb{R}^{d_k \times d_{k+1}}$ and $\mW^{(k)}_{\rm neigh} \in \mathbb{R}^{d_k \times d_{k+1}}$ are trainable parameter matrices and $\sigma(\cdot)$ is an elementwise non-linearity (e.g., a \texttt{ReLU}). Here, the graph operator $f^{(k)}(\cdot)$ that performs neighborhood aggregation is actually a first-order graph filter.

For graph property prediction or classification, given a set of graphs $\{\mathcal{G}_1, \cdots, \mathcal{G}_M\}$ and their labels $\{y_1,\cdots,y_M\}$, the representation vector ${\boldsymbol \varepsilon}_G$ of an entire graph $\mathcal{G}$ is required to predict its label $y_G = \texttt{PREDICT}({\boldsymbol \varepsilon}_G)$, where  $\texttt{PREDICT}(\cdot)$ is a trained decoder. The graph embedding ${\boldsymbol \varepsilon}_G$ is computed using a node-level pooling or readout function that operates on the node representation vectors as
\begin{equation}
	{\boldsymbol \varepsilon}_G = \texttt{READOUT}\left(\left\{\vx_1^{(K)},\cdots, \vx_N^{(K)}\right\}\right),
	\label{eq:SAGNNs_readout}
\end{equation}
where typical choices for the readout function $\texttt{READOUT}(\cdot)$ are concatenation, summation, mean, max~\cite{JK,SAG}, hierarchical pooling~\cite{diffpool}, and sort pooling~\cite{DGCNN}. 

\subsection{Weisfeiler-Lehman isomorphism test} Determining whether two graphs are isomorphic is a difficult problem with no known polynomial time solution~\cite{garey}. The Weisfeiler-Lehman (WL) vertex refinement algorithm~\cite{babai} produces for each graph under test a canonical form, which when not equivalent implies that the graphs are not isomorphic. However, for non-isomorphic graphs that lead to the same canonical form, the WL test is not useful in distinguishing the graphs under test. The WL vertex refinement algorithm (also referred to as the one-dimensional WL or 1-WL test) iteratively updates the labels of a node based on the labels of its neighboring nodes and assigns a unique label. For a graph~$\mathcal{G}$, 1-WL vertex refinement iteration is given as
\begin{eqnarray}
\label{eq:WL}
	l^{(t+1)}_v = \varphi\left(l^{(t)}_v,\left\{l^{(t)}_ i,\; \forall\; i \in \mathcal{N}_v^{(1)}\right\}\right),
\end{eqnarray}
where $l^{(t)}_v$ is the unique label for node $v \in \mathcal{V}$ at the $t$-th iteration. Here, $\mathfrak{L}_v = \left\{l^{(t)}_i,\; \forall\; i \in \mathcal{N}_v^{(1)}\right\}$ is a multiset of the neighborhood labels for node $v$ as different nodes can have identical labels and $\varphi(\cdot): \mathfrak{L}_v \mapsto \mathbb{R}$ is an injective hashing function that maps a multiset of different neighborhood labels to a distinct label. The iterative procedure in Equation~\eqref{eq:WL} is simultaneously applied on two graphs under test to refine the labels until convergence. 

A GNN model is said to be as powerful as the 1-WL test if it generates different graph embeddings for two graphs, identified as non-isomorphic by the 1-WL test. The 1-WL vertex refinement iterations in \eqref{eq:WL} are similar in nature to the feature update iterations of SA-GNNs in~\eqref{eq:SA-GCN}. This similarity has been leveraged to theoretically characterize the discriminative power of some of the popular GNN variants, such as GCN~\cite{GCN}, GraphSAGE~\cite{GraphSAGE}, and to build  GIN~\cite{GIN}, which is as powerful as the 1-WL test. 

\section{GNNs with parallel neighborhood aggregation}\label{sec:arch}
In this section, we present a generic PA-GNN model based on parallel neighborhood aggregations. The model presented in this section is generic as we do not restrict it to a specific aggregation or readout procedure. We also provide theoretical conditions under which the presented model is provably as powerful as the 1-WL test.

\subsection{A generic PA-GNN model} \label{sec:generic_model}

The representation vector of a node capturing the structural information at multiple scales related to different neighborhood depths can be simultaneously obtained by choosing an appropriate neighborhood aggregation graph operator. These node embeddings from different depths are then combined using readout functions to obtain graph embeddings. Formally, the proposed generic PA-GNN model with $R+1$ branches has the following three main components.

\textbf{Parallel neighborhood aggregation}: At the $r$th branch, the intermediate representation vector of node $v$ is computed by aggregating feature vectors from its $r$-hop neighbor nodes as
\begin{align}
   \vz_v^{(r)} &= g^{(r)}\left(f_v^{(r)}\left(\left\{\vx_j, \;\forall\; j\in \mathcal{N}_v^{(r)}\right\} \right)\right),
   \label{eq:SPIN_local_agg}
\end{align}
where $f_v^{(r)}(\cdot)$ is a graph operator that does neighborhood aggregation and $g^{(r)}(\cdot): \mathbb{R}^d\mapsto \mathbb{R}^{{d}'}$ is a learnable transformation function. The updated node features  $f_v^{(r)}\left(\left\{\vx_j, \;\forall\; j\in \mathcal{N}_v^{(r)}\right)\right\}$ can be efficiently precomputed  outside the training process as they do not depend on the learnable parameters. 

\noindent\textbf{Branch-level graph pooling}:
At the $r$th branch, we have a pooling function that maps a set of node embeddings at each branch to an intermediate graph embedding as
\begin{equation}
    	\vs^{(r)}_G = \Omega \left(\left\{\vz_v^{(r)}, \; \forall v \in \mathcal{V}\right\}\right) 
	\label{eq:SPIN_branchlevel}, \,\,\, r= 0,1,\ldots,R,
\end{equation}
where $\Omega(\cdot)$ is a readout function.

\noindent\textbf{Global readout}: Finally, we have a global readout function that combines graph embeddings from each branch to obtain the final graph embedding as
\begin{equation}
	\ve_G = \psi\left(\left\{\vs^{(0)}_G,\cdots, \vs^{(R)}_G\right\}\right) \label{eq:SPIN_graphlevel},
\end{equation}
where $\psi(\cdot)$ is the global readout function.

For node and link prediction tasks, one may compute the final representation vector of node $v$ by combining the transformed intermediate representation vectors from the $R+1$ branches as
\begin{equation}
    \vh_v = \Theta \left(\vz_v^{(0)},\vz_v^{(1)},\cdots,\vz_v^{(R)}\right),
    \label{eq:SPIN_noderep}
\end{equation}
where $\Theta(\cdot)$ is the learnable global aggregation function. The branch-level and graph-level readouts are usually not required for node and link prediction tasks.

In what follows, we provide conditions for the functions $f_v^{(r)}(\cdot)$, $g_v^{(r)}(\cdot)$, $\Omega(\cdot)$, and $\psi(\cdot)$ to theoretically characterize the discriminative power of a PA-GNN model.

\subsection{Theoretical characterization}
PA-GNN models of the form described in Section~\ref{sec:generic_model} are fundamentally different from SA-GNN models as PA-GNN models reduce to standard neural networks as the neighborhood aggregations are not iterative and precomputed. More importantly, the global readout in Equation~\eqref{eq:SPIN_graphlevel} combines graph embeddings (unlike, the readout in Equation~\eqref{eq:SAGNNs_readout}). Therefore, in this work, we extend the theoretical framework for analyzing and characterizing the discriminative power of SA-GNN models in~\cite{GIN} to PA-GNN models. 

Our next theorem states conditions required for PA-GNNs to be powerful as the 1-WL test.
\begin{theorem} \label{theo:power}
\textit{A PA-GNN model $\mathcal{P}$ with $R+1$ branches maps two non-isomorphic graphs $\mathcal{G}_1$ and $\mathcal{G}_2$ as identified by the 1-WL test to two different embeddings if:}
\begin{enumerate}
    \item $\mathcal{P}$ produces node vector representations according to Equation~\eqref{eq:SPIN_local_agg} with injective functions $g^{(r)}(\cdot)$ and $f_v^{(r)}(\cdot)$.

    \item The branch-level and global readout functions $\Omega(\cdot)$ and $\psi(\cdot)$, in Equations~\eqref{eq:SPIN_branchlevel} and \eqref{eq:SPIN_graphlevel}, respectively, are also injective.
\end{enumerate}
\end{theorem}

We prove Theorem~\ref{theo:power} in Appendix~\ref{app:theo1}. The proof extends the setting in~\cite{GIN} from iterative feature vector updates to simultaneous and parallel feature vector computations. To do so, we leverage the fact that the precomputed local aggregation at the $r$-th branch can be realized by successive local aggregations and that the composition of injective multiset functions is injective. We then show via mathematical induction that there always exists an injective function $\tau(\cdot)$ such that $\mathbf{h}_v = \tau(l_v^{(R)})$, where $l_v^{(R)}$ is the label generated for node $v$ at the $R$-th iteration by the WL vertex refinement algorithm [cf. Equation~\eqref{eq:WL}] and $\mathbf{h}_v$ is representation vector of node $v$ produced by a PA-GNN model with $R+1$ branches [cf. Equation~\eqref{eq:SPIN_noderep}]. Thus choosing functions in Equations~\eqref{eq:SPIN_local_agg}-\eqref{eq:SPIN_graphlevel} appropriately, we obtain a PA-GNN model that is provably as powerful as the 1-WL test. We next build one such PA-GNN model, which we call {\it simple and parallel graph isomorphism network} (SPIN). 

\section{Simple and parallel graph isomorphism network (SPIN)} \label{sec:SPIN}

\begin{figure*}[!t]
    \centering
    \includegraphics[width=1\columnwidth]{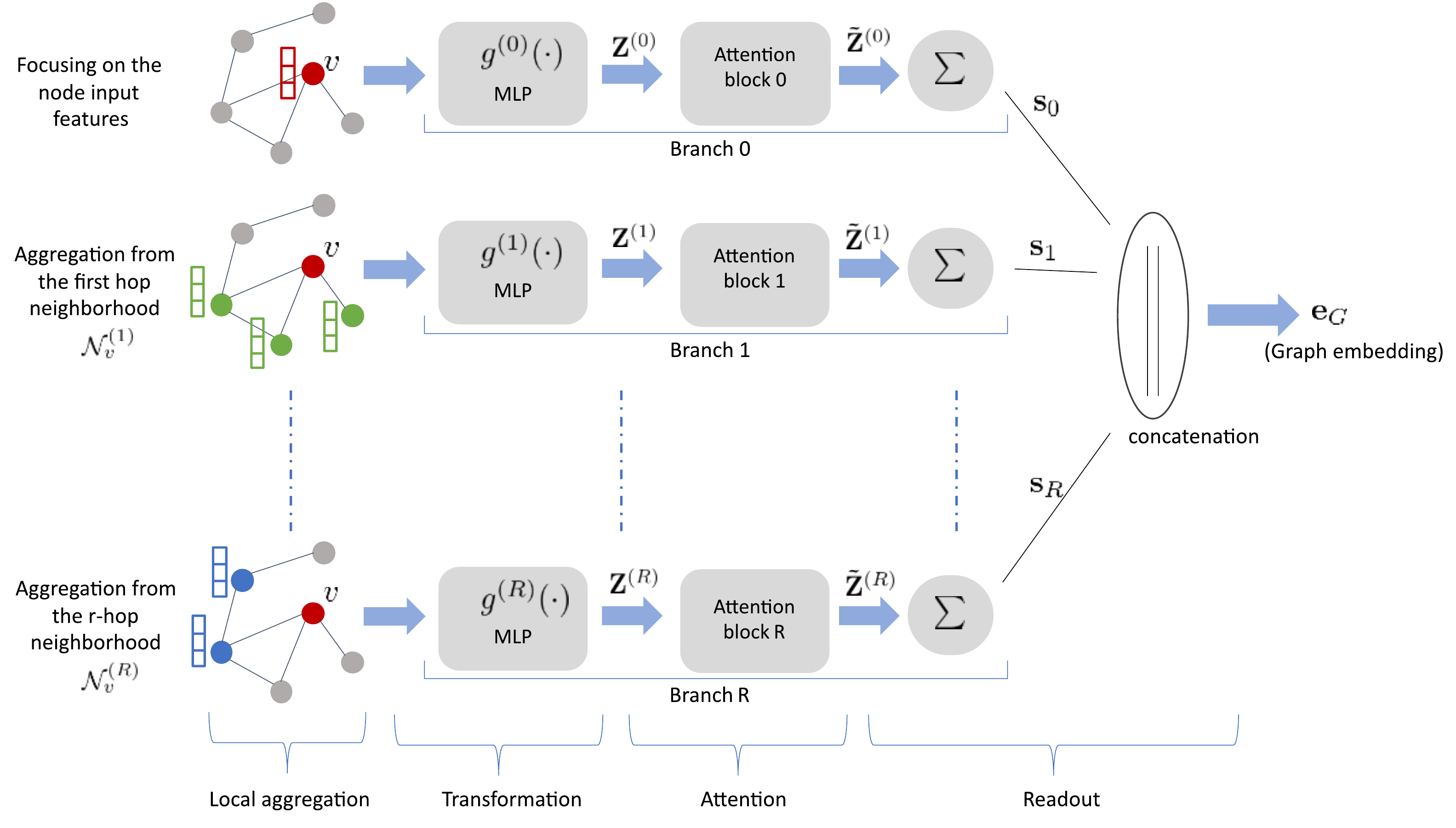}
    \caption{The SPIN architecture.}
    \label{fig:arch}
\end{figure*}

SPIN specializes functions in the generic PA-GNN model so that the conditions provided in Theorem~\ref{theo:power} are satisfied. Specifically, SPIN has three main components: parallel neighborhood aggregations, branch-level readouts, and a global readout as illustrated in Figure~\ref{fig:arch}. 

\subsection{Parallel neighborhood aggregations} Let us collect the neighborhood feature vectors of node $v$ at the $r$-th branch in  $\mathfrak{X}_v^{(r)} = \{\vx_j,\; \forall j \in \mathcal{N}_v^{(r)}\}$, which is a multiset. Then the neighborhood aggregation operator at the $r$th branch $f_v^{(r)}: \mathfrak{X}_v^{(r)} \mapsto \mathbb{R}^d$ is a multiset function. For $f_v^{(r)}$, we use integer powers of the graph operator. Let us define the updated feature vector as
\[
\mB^{(r)} = [\vb_1^{(r)},\cdots,\vb_N^{(r)}]\rT = \bar{\mA}^r\mX,
\]
where the graph filtering operation $\bar{\mA}^r\mX$ implements a sum of the $r$-hop neighbor node embeddings. In other words, for $f_v^{(r)}$, we use integer powers of the graph operator as $f_v^{(r)}\left(\mathfrak{X}_v^{(r)}\right) = \vb_v^{(r)}.$

Some choices of the graph operator $\bar{\mA}$ are the adjacency matrix $\mA$, which is the basic sum aggregator and is injective, the normalized adjacency matrix $\tilde{\mA} = \mD^{-1/2}\mA\mD^{-1/2}$ with the diagonal degree matrix $\mD$ or its linear combination $\bar{\mA} = \tilde{\mA}+\mA$. When the pair of graphs under test are regular and are of the same size but different node degrees, degree normalized graph operators may lead to the same node embeddings, hence not injective and less powerful as the 1-WL test~\cite{GAMLP, Morris}. Although such cases are rare in practice, when encountered (easily verified via prescreening the graphs to be tested), we can use $\bar{\mA} = \tilde{\mA}+\mA$, where $\mA$ retains the degree information and $\tilde{\mA}$ provides the advantages of normalization~\cite{GCN}.

\subsection{Branch-level readouts} In a PA-GNN model with $R+1$ branches, we have representation vectors of all the nodes in the graph at each branch. We combine these node representation vectors to compute $R+1$ branch-level graph embeddings $\vs_G^{(r)}$ for~$r=0,\cdots, R$. At the $r$th branch, this amounts to pooling the $r$-hop representation vectors of all the nodes in the graph. We perform this pooling through a weighted summation as
\[
\vs_G^{(r)} = \sum_{v \in \mathcal{V}}\alpha_v^{(r)} g^{(r)}\left(\vb_v^{(r)}\right) = \sum_{v \in \mathcal{V}}\alpha_v^{(r)} \textbf{z}_v^{(r)}, 
\]
where the local features $\vb_v^{(r)} = f_v^{(r)}(\mathfrak{X}_v^{(k)})$ are first transformed as $\vz_v^{(r)} = g^{(r)}(\vb_v^{(r)})$. The weights $\alpha_v^{(r)}$ may be used to focus on the most relevant local features (discussed later on). Lemma~\ref{lem:MLP} suggests that modeling transformation functions $g^{(r)}(\cdot)$ as single-layer perceptrons (SLPs) cannot distinguish multiset local neighborhood features. 

\begin{lemma} \label{lem:MLP}
\emph{For two distinct multisets $\mathfrak{X}_1 \neq \mathfrak{X}_2$, the weighted summation of their linear mappings can be equal with ReLU (or leaky ReLU) as a non-linearity, i.e., $\sum_{\vx\in \mathfrak{X}_1}\alpha_i \texttt{ReLU}(\mW\vx) = \sum_{\vx\in \mathfrak{X}_2}\beta_i \texttt{ReLU}(\mW\vx)$, for any $\mW$.}
\end{lemma}
We prove the lemma in the Appendix~\ref{app:lemma1}. This lemma is a generalization of~\cite[Lemma 7]{GIN} to the case with a weighted sum that allows us to include an attention mechanism as discussed next. Thus we use multi-layer perceptrons (MLPs) to model $g^{(r)}(\cdot)$.

\begin{figure*}[!t]
    \centering
    \includegraphics[width=1\linewidth]{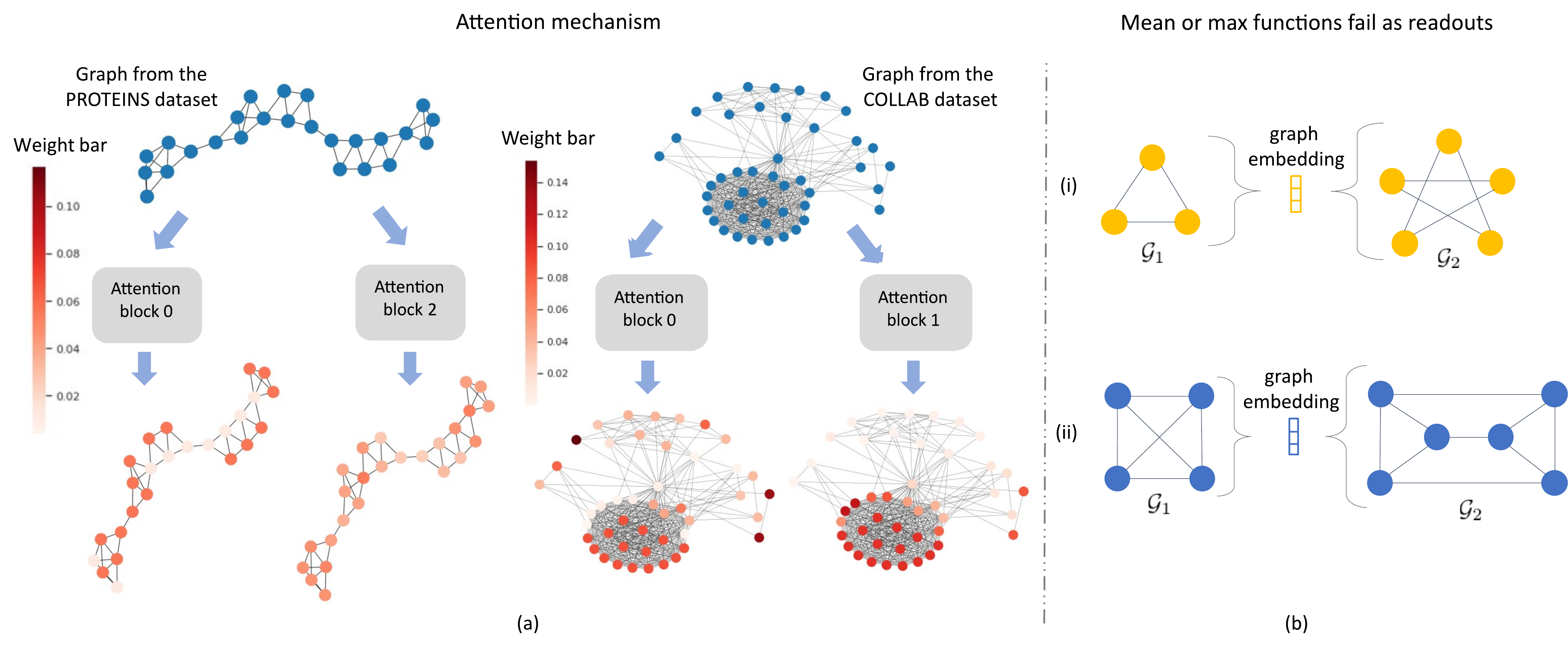}
    \caption{(a) Illustration of the attention mechanism. Node color indicates its weights. Darker the shade, more the weight. (b) Examples of graph structures that the mean and max readout functions fail to distinguish.}
    \label{fig:readout_att}
\end{figure*}

Inspired from the attention mechanism~\cite{GAT} for pooling, i.e., self-attention pooling~\cite{SAG}, we design weights through an attention mechanism to focus on the most relevant parts of the local features and retain the most important ones. Formally, the weight of node $v$ at the $r$-th branch is computed as
\begin{align}    
    \alpha_v^{(r)} &= \texttt{ATTENTION}\left(\vw^{(r)}, \vz_v^{(r)}\right) = \frac{\exp(\beta_v^{(r)})}{\sum_{u \in \mathcal{V}} \exp(\beta_u^{(r)})}
\label{eq:att_1}
\end{align}
with $\beta_v^{(r)} = \texttt{ReLU}\left(\langle\vw^{(r)}, \vz_v^{(r)}\rangle\right)$. Here, $\vw^{(r)}$ is the learnable vector that extracts the attention weight $\alpha_v^{(r)}$ for node $v$ at the $r$-th branch based on its local feature vector $\vz_v^{(r)}$. For instance, $\textbf{z}_v^{(r)}$ carries the information about the $r$-hop neighborhood for node $v$, and $\alpha_v^{(r)}$ tells how significant that information is in comparison with the $r$-hop information of the other nodes. Our next lemma states that the attention mechanism preserves injectivity.

\begin{lemma} \label{lem:selfattention}
\emph{For two distinct features $\vz_1$ and $\vz_2$, $\alpha_1 \vz_1$ and $\alpha_2 \vz_2$  are also distinct, if $\alpha_i = \texttt{ATTENTION}(\vw, \vz_i)$ for $i=1,2$, and for any $\vw$.}
\end{lemma}

We prove the lemma in the Appendix~\ref{app:lemma2}. Figure~\ref{fig:readout_att}(a) illustrates the weights produced by the attention block at branches $r=0$, $r=1$, or $r=2$ for sample graphs from the COLLAB~\cite{social} and the PROTEINS dataset~\cite{proteins}. We can observe that the same nodes are given different weights in different branches. When attention mechanism is not used, the weights $\alpha_v^{(r)}$ are all set to 1. 

Finally, we need to combine the representation vectors of all the nodes in the graph to arrive at a branch-level graph representation vector. Usually used readout operators to compute the representation vector for an entire graph are the mean, max, or a combination of max and mean readout functions~\cite{JK}. Our next lemma states that mean or max graph-level readout functions fail to distinguish graph structures. 

\begin{lemma} \label{lemm:readout}
\emph{Consider two undirected graphs $\mathcal{G}_1$ and $\mathcal{G}_2$ with different number of nodes and structural connectivity. The embeddings 
$\ve_{\mathcal{G}_1}$ and $\ve_{\mathcal{G}_2}$ of the entire graphs $\mathcal{G}_1$ and $\mathcal{G}_2$, respectively, from mean or max graph-level readout functions are equal if both the graphs are regular and have the same type of nodes (i.e., they have the same feature vectors).}
\end{lemma}

We prove the lemma in Appendix~\ref{app:lemma3}. A similar result that the mean and max node-level aggregators to sequentially update the node features are less powerful was provided~\cite{GIN}. The main difference is that the mean and max functions here are used to obtain the graph embeddings. Some example graphs that cannot be discriminated using mean or max readouts are shown in Fig.~\ref{fig:readout_att}(b), where $\mathcal{G}_1$ and $\mathcal{G}_2$ are regular and have the same type of nodes (indicated by the same color). Therefore, we use the basic summation operator $\sum_{v \in \mathcal{V}} \tilde{\vz}_v^{(r)}$ that preserves injectivity to obtain the branch-level graph embedding. This, in other words, means that each branch of SPIN is as powerful as the 1-WL test. 

\subsection{Global readout} Global readout refers to the pooling of branch-level graph embeddings to obtain a single representation for an entire graph. 
From Lemma~\ref{lemm:readout}, to retain the discriminative power of the branch-level embeddings, we use injectivity preserving concatenation (and not mean or max) function to perform a global readout to obtain the representation vector of an entire graph as
\begin{eqnarray}
\label{eq:global}
\ve_G = \psi\left(\left\{\vs^{(0)},\cdots, \vs^{(R)}\right\}\right) = \left[\vs^{(0)T},\cdots, \vs^{(R)T}\right]\rT,
\end{eqnarray}
where $\psi(\cdot)$ is the multiset concatenation function. 

This completes the SPIN architecture, which satisfies the conditions provided in Theorem~\ref{theo:power} and thus is as powerful as the 1-WL test. 

\subsection{Computational complexity}
The time complexity of GCN (a SA-GNN model) with $L$ graph convolution layers is about ${\mathcal{O}}(L( Nd^2 + |\mathcal{E}|d))$ for a graph $\mathcal{G}=(\mathcal{V,E})$ having $|\mathcal{E}|$ edges and $N$ nodes with each node having a $d$-dimensional feature vector~\cite{survey}, where for simplicity, we retain $d$-dimensional features in all the hidden layers. Here, the $Nd^2$ term corresponds to the computations involved in the feature transformation, whereas the $|\mathcal{E}|d$ term is due to the neighborhood aggregation performed during training. GIN has a similar complexity as GCN, as it also does a sequential neighborhood aggregation. Although GraphSAGE performs neighborhood sampling to reduce the computation cost, the neighbor nodes are identified during training. The number of edges $|\mathcal{E}|$ increases the compute requirements of SA-GNN models, particularly for dense graphs, where $|\mathcal{E}| \propto N^2$. In contrast, a PA-GNN model performs neighborhood aggregations beforehand, due to which its runtime is independent of the number of edges in graphs. Specifically, a PA-GNN model with $R+1$ branches has a time complexity of $\mathcal{O}(RNd^2)$. The time complexity due to branch-level and global readouts does not depend on the size and structure of the graph.

\section{Experiments}\label{sec:exp}
This section describes the experiments performed to evaluate SPIN and compare its graph classification accuracy with state-of-the-art GNN variants.

\begin{table*}[!t]
    \centering
    \caption{\MakeUppercase{Testing accuracy on the social domain datasets. NA indicates unavailability of scores due to their high computational requirement. Top performance are highlighted as} \textcolor{red}{FIRST} AND THE \textcolor{blue}{SECOND}. BEST SEEN IN COLOR.}
    \label{Tab: social_results}
    \scalebox{0.925}{
    \begin{tabular}{*6c}
        \hline
        {}   & IMDB-BINARY & IMDB-MULTI & REDDIT-BINARY & REDDIT-MULTI & COLLAB\\
        \hline
        Baseline & $70.8\pm 5.0$ & $49.1\pm 3.5$ & $82.2\pm 3.0$ & $52.2\pm 1.5$ & $70.2\pm 1.5$\\
        DGCNN & $69.2\pm 3.0$ & $45.6\pm 3.4$ & $87.8\pm 2.5$ & $49.2\pm 1.2$ & $71.2\pm 1.9$\\
        DiffPool & $68.4\pm 3.3$ & $45.6\pm 3.4$ & $89.1\pm 1.6$ & $53.8\pm 1.4$ & $68.9\pm 2.0$\\
        ECC & $67.7\pm 2.8$ & $43.5\pm 3.1$ & NA & NA & NA\\
        GIN & \textcolor{blue}{$\mathbf{71.2\pm 3.9}$} & $48.5\pm 3.3$ & \textcolor{blue}{$\mathbf{89.9\pm 1.9}$} & \textcolor{red}{$\mathbf{56.1\pm 1.7}$} & \textcolor{red}{$\mathbf{75.6\pm 2.3}$}\\
        GraphSAGE & $68.8\pm 4.5$ & $47.6\pm 3.5$ & $84.3\pm 1.9$ & $50.0\pm 1.3$ & \textcolor{blue}{$\mathbf{73.9\pm 1.7}$}\\
        \hline
        SPIN-att & \textcolor{red}{$\mathbf{74.5\pm 3.3}$} & \textcolor{red}{$\mathbf{50.4\pm 4.1}$} & \textcolor{red}{$\mathbf{90.2\pm 1.6}$} & \textcolor{blue}{$\mathbf{54.3\pm 1.6}$} & $72.0\pm 1.7$\\
        SPIN-non-att & $71.1\pm 5.0$ & \textcolor{blue}{$\mathbf{50.1\pm 4.2}$} & $88.4\pm 2.5$ & $53.8\pm 1.4$ & $71.6\pm 2.1$\\
        \hline
    \end{tabular}}
\end{table*}

\begin{table*}[!t]
    \centering
      \caption{\MakeUppercase{Testing accuracy on the chemical domain datasets. For the OGBG-MOLHIV dataset, we report AUROC. Top performance are highlighted as} \textcolor{red}{FIRST} AND THE \textcolor{blue}{SECOND}. BEST SEEN IN COLOR.}
      \label{Tab: bio_results}
        \begin{tabular}{*6c}
        \hline
        {}   & D\&D & NCI1 & PROTEINS & ENZYMES & OGBG-MOLHIV\\
        \hline
        Baseline & \textcolor{red}{\textbf{78.4 $\pm$ 4.5}} & 69.8$\pm$ 2.2 & \textcolor{red}{\textbf{75.8$\pm$ 3.7}} & \textcolor{red}{$\mathbf{65.2\pm 6.4}$} & $0.708 \pm 0.02$\\
        DGCNN & $76.6\pm 4.3$ & $76.4\pm 1.7$ & $72.9\pm 3.5$ & $38.9\pm 5.7$ & $0.722 \pm 0.03$ \\
        DiffPool & $75.0\pm 3.5$ & \textcolor{blue}{$\mathbf{76.9\pm 1.9}$} & $73.7\pm 3.5$ & $59.5\pm 5.6$ & $0.753 \pm 0.03$ \\
        ECC & $72.6\pm 4.1$ & $76.2\pm 1.4$ & $72.3\pm 3.4$ & $29.5\pm 8.2$ & NA \\
        GIN & $75.3\pm 2.9$ & \textcolor{red}{$\mathbf{80.0\pm 1.4}$} & $73.3\pm 4.0$ & \textcolor{blue}{$\mathbf{59.6\pm 4.5}$} & \textcolor{red}{$\mathbf{0.768 \pm 0.03}$} \\
        GraphSAGE & $72.9\pm 2.0$ & $76.0\pm 1.8$ & $73.0\pm 4.5$ & $58.2\pm 6.0$ & $0.757 \pm 0.02$\\
        \hline
        SPIN-att & $72.8\pm 4.0$ & $73.5\pm 1.8$ & $73.9\pm 3.6$ & $54.9\pm 5.2$ & $0.729 \pm 0.02$ \\
        SPIN-non-att & \textcolor{blue}{$\mathbf{78.3\pm 3.8}$} & $74.0\pm 1.7$ & \textcolor{blue}{$\mathbf{75.6\pm 4.5}$} & $53.5\pm 5.7$ & \textcolor{blue}{$\mathbf{0.761\pm 0.02}$}\\
        \hline
    \end{tabular}
\end{table*}

\begin{table}[!t]
        \caption{\MakeUppercase{Testing accuracy on the brain datasets. Top performance are highlighted as} \textcolor{red}{FIRST} AND THE \textcolor{blue}{SECOND}. BEST SEEN IN COLOR.}
        \label{Tab: brain_results}
        \centering
        \begin{tabular}{*3c}
        \hline
        {}   & OHSU & PEKING-1 \\
        \hline
        Baseline & $58.9 \pm 9.5$ & $59.1 \pm 3.6$\\
        DGCNN & $60.2 \pm 6.7$ & $62.1 \pm 4.3$\\
        DiffPool &	$61.9 \pm 7.8$ &	$63.1 \pm 5.2$\\
        ECC & $61.4 \pm 5.9$ &	$61.1 \pm 4.2$\\
        GIN & \textcolor{red}{$\mathbf{65.0 \pm 5.6}$} &	\textcolor{blue}{$\mathbf{63.4 \pm 5.3}$}\\
        GraphSAGE &	$62.5 \pm 7.7$&	$62.1 \pm 5.7$\\
        \hline
        SPIN-att &	$61.9 \pm 5.5$&	\textcolor{red}{$\mathbf{64.2 \pm 6.3}$}\\
        SPIN-non-att &	\textcolor{blue}{$\mathbf{63.2 \pm 7.7}$} &	$62.3 \pm 4.8$\\
        \hline
    \end{tabular}
\end{table}

We evaluate SPIN on twelve diverse benchmark datasets for binary as well as multiclass graph classification tasks. Specifically, we evaluate on five chemical domain datasets: D\&D~\cite{DD}, PROTEINS~\cite{proteins}, NCI1~\cite{nci1}, ENZYMES~\cite{enzymes} and OGBG-MOLHIV~\cite{ogb}, five social domain datasets: IMDB-BINARY, IMDB-MULTI, COLLAB, REDDIT-BINARY and REDDIT-MULTI~\cite{social}, and two brain network datasets: OHSU and PEKING-1~\cite{brain}. All the datasets are publicly available~\cite{public,brain,ogb} and are commonly used for evaluating GNNs. More details about these datasets are provided in Table~\ref{Tab:stats} at the end of the paper. 

\subsection{Experimental setup}

We follow the evaluation procedure described in~\cite{Fair}, which suggests a standard procedure for evaluating machine learning models for graph classification. Based on the sources of dataset curation, the social and the chemical datasets (except OGBG-MOLHIV) are collectively referred to as the TU datasets, OGBG-MOLHIV as the open graph benchmark (OGB) dataset. We next describe the experimental setup for each of them.

\subsubsection{TU datasets} For all experiments with TU datasets, we use input features as suggested by~\cite{Fair}. Specifically, for molecular graphs from the chemical domain, the nodes are augmented with the one-hot encodings of their atom types. For the ENZYMES dataset, we append the available 18-dimensional node attributes with the one-hot encodings of their atom types. We conduct experiments on the predefined stratified data splits provided by~\cite{Fair}. We use an inner holdout technique with 90\%-10\% training-validation split, and each selected model is trained three times on a testing fold to eliminate any weight initialization biases. We use 10-fold cross-validation and report the average testing accuracy of all the folds with its standard deviation in Tables~\ref{Tab: social_results} and~\ref{Tab: bio_results}. 
\subsubsection{OGB dataset} We consider a molecular graph prediction dataset, referred to as OGBG-MOLHIV~\cite{moleculenet}. This is a large-scale dataset with 41127 graphs to be classified. The nodes in the graph are augmented with the available 9-dimensional input node attributes. We use the same scaffold data split for evaluation as in~\cite{ogb}. As the dataset is highly skewed, we report the area under the ROC (AUROC) score for the testing set. The process is repeated thrice to avoid any weight initialization biases, and the standard deviation is reported in Table~\ref{Tab: bio_results}.

\subsubsection{Brain datasets} We also consider two brain datasets, namely, OHSU and PEKING-1, which are datasets used for hyper/impulsive and gender classification studies, respectively~\cite{brain}. Graphs are constructed using a CC200 parcellation on the brain functional magnetic resonance image (fMRI) data by mapping each region of the brain as a node and modeling the similarity between these regions through the edges. Similar to the social domain datasets, we use the nodal degree information as input features for these brain datasets. We use a 90\%-10\% training-validation split for model selection and train each selected model for three times to avoid the initialization biases. We do a 5-fold cross-validation and report the average testing accuracy and the standard deviation across all the folds in Table~\ref{Tab: brain_results}. 

\subsection{Training and baselines}

For all the datasets, we train SPIN in a supervised manner using the cross-entropy loss function, which for a minibatch $\mathcal{B} = \{\mathcal{G}_1,\mathcal{G}_2,\cdots,\mathcal{G}_M\}$ of $M$ graphs and a $K$-class classification problem, is defined as
\begin{eqnarray}
\label{eq:loss}
\mathcal{L} = -\frac{1}{M}\sum_{\mathcal{G}_i\in \mathcal{B}}\sum_{k=1}^{K}y_{ik}\log(\texttt{softmax}(\texttt{MLP}(\ve_{G_i}))),
\end{eqnarray}
where $\{y_{i1},y_{i2},\cdots,y_{iK}\}$ are the one-hot labels associated with graph $\mathcal{G}_i$. The graph label is predicted using an $\texttt{MLP}$ classifier with the graph embedding $\ve_{G_i}$ as input so that $\texttt{MLP}(\ve_{G_i}) \in \mathbb{R}^K$. The components of this $K$-dimensional output vector with softmax function are interpreted as the probabilities assigned to the $K$ classes.
We perform experiments on SPIN with attention mechanism (referred to as SPIN-att) and without attention (referred to as SPIN-non-att) obtained by setting $\alpha_v^{(r)} = 1$  to validate the performance gain achieved using the attention mechanism. We implement early stopping while training, i.e., terminate the training process if there is no significant gain in the validation accuracy after a certain number of training epochs, and use standard techniques like the L2 regularization, dropout, and batch normalization to avoid overfitting. All the hyperparameters, namely, batch size, learning rate, intermediate node representation vector dimensions, depth of a GNN model, L2 regularization parameters, are summarized in Table~\ref{Tab:hyper} at the end of the paper. 

We compare the graph classification performance of SPIN with non-graph methods, indicated as \textit{Baseline} in Tables~\ref{Tab: social_results},~\ref{Tab: bio_results}, and~\ref{Tab: brain_results}. Molecular fingerprint technique~\cite{MFT1,MFT2} is used as a baseline for the chemical datasets, while (non-graph) multi-layer perceptrons~\cite{deepsets} are used for the ENZYMES, social domain, and brain datasets. Such a comparison with non-graph methods reveals the capability of GNNs in exploiting the graph topology. Furthermore, we compare SPIN with GNN variants commonly used for graph classification, namely, DGCNN~\cite{DGCNN}, DiffPool~\cite{diffpool}, ECC~\cite{ECC}, GIN~\cite{GIN}, and GraphSAGE~\cite{GraphSAGE}.

\subsection{Results and discussion}
 
Tables~\ref{Tab: social_results},~\ref{Tab: bio_results}, and~\ref{Tab: brain_results} report the performance of SPIN on the social domain, chemical domain, and brain datasets, respectively. Firstly, SPIN can capture and exploit the underlying graph structure, which can be seen from the results as it outperforms the baseline on all the social domain, brain datasets and NCI1 from the chemical domain. On D\&D, PROTEINS, and ENZYMES datasets, no GNN model exceeds the baseline, suggesting that the structural information is not important for these datasets. Imposing an inductive bias to learn the graph-structural information for these three datasets deteriorates the model's performance. Interestingly, on the D\&D and PROTEINS datasets, SPIN outperforms other GNN variants and is on par with the baseline, demonstrating its selective nature of utilizing the topological information whenever needed. Furthermore, SPIN-att outperforms SPIN-non-att on all the social domain and PEKING-1 from brain datasets. This might be due to the degree information encoded as the input features for the social domain datasets that aid in identifying and attending the relevant nodes for the classification task at hand. In the chemical domain datasets, all the nodes play an important role, representing a particular type of atom; see in Fig.~\ref{fig:readout_att}(a) that all the nodes in a sample graph from the PROTEINS dataset have similar attention weights. Suppressing features of an atom from a molecular graph has a negative impact in predicting the type of that molecule, as can be seen from the results. On the D\&D and PROTEINS datasets, SPIN-non-att outperforms SPIN-att and other GNN variants. 

In summary, SPIN performs competitively or better than existing GNNs on the benchmark datasets while incurring less training and inference time as the training runtime does not depend on the graph structure (the number of edges).

\section{Conclusions}\label{sec:conclusions}
We have presented a theoretical framework to characterize and analyze the discriminative power of PA-GNN models, which are GNNs that perform neighborhood aggregation in parallel and before the training procedure. Consequently, PA-GNN models are independent of the graph structure in terms of computations, but well-capture the graph structure. We have provided conditions under which PA-GNN models are provably as powerful as the 1-WL test. Although the node embedding aggregation in PA-GNN models has apparently a different form than the iterative label update procedure of the 1-WL algorithm, we have shown that the node labels from the 1-WL test are injectively related to the node embeddings generated by a PA-GNN model. We have also presented an example GNN model, namely, SPIN, that obeys the prescribed theoretical conditions. We have demonstrated via experiments that SPIN outperforms state-of-the-art methods on a majority of graph classification benchmark datasets related to social, chemical, and brain networks.

\section*{Appendix}
\appendix
\section{Proof for Theorem 1} \label{app:theo1}
We prove that a PA-GNN model that satisfies the conditions provided in Theorem 1 is as powerful as the 1-WL test. Unlike the iterative node embedding update procedure in SA-GNN models, the procedure to compute the node embeddings in PA-GNN models has apparently a different form than the iterative label update procedure in the 1-WL vertex refinement algorithm. In what follows, we show that the node labels generated from the 1-WL algorithm can be mapped to the node embedding generated by a PA-GNN model through an injective function. The proof has two parts. In the first part, we relate the embedding of a node from a PA-GNN model to its label from the 1-WL algorithm. In the second part, we relate the graph embeddings from a PA-GNN model to the 1-WL test. 

For a graph $\mathcal{G} = (\mathcal{V,E})$, recall that the 1-WL vertex refinement update equation for a node $v \in \mathcal{V}$ is given by
\begin{align*}
    l^{(t+1)}_v = \varphi\left(l^{(t)}_v,\left\{l^{(t)}_j, \;\forall j\in \mathcal{N}_v^{(1)}\right\}\right),
\end{align*} 
where $\mathcal{N}_v^{(1)}$ is the set of one-hop neighbors of node $v$. The 1-WL iteration is initialized with the input features as $l_v^{(0)} = \vx_v, \; \;\forall v \in \mathcal{V}$. Also, recall that the node embeddings generated using a PA-GNN model with $R$ branches are given by
\begin{align*}
    \vh_v &= \Theta\left(\vz_v^{(0)},\vz_v^{(1)},\cdots,\vz_v^{(R)}\right)
\end{align*}
with    $\vz_v^{(r)} = g^{(r)}\left(f_v^{(r)}\left(\left\{\vx_j,\; \forall  j \in \mathcal{N}_v^{(r)}\right\}\right)\right).$ 

Let us introduce the embedding of node $v$, $\vh_v^{(r)}$, that gathers information from its $r$-hop neighborhood (i.e., from $r$ branches), and define it as $\vh_v^{(r)} = \Theta_r\left(\vz_v^{(0)},\cdots,\vz_v^{(r)}\right)$ with $\Theta_{r}(\cdot)$ being the transformation function that operates on the $r$ intermediate embeddings. Let us also introduce the composite function $c_v^{(r)} = g^{(r)}\circ f_v^{(r)}$. Here, the notation $ g \circ f$ means that the function $g$ is composed with the function $f$. We frequently use the fact that \textit{composition preserves injectivity}.

\subsection{Relating the node embeddings in a PA-GNN model to the labels in the 1-WL algorithm}
To begin with, we analyze the relation between the node embeddings $\vh_v^{(r)}$ generated non-iteratively using a PA-GNN model and the labels $l_v^{(r)}$ generated by the iterative 1-WL algorithm. Through induction, we show that there always exists a function $\tau_r(\cdot)$ such that $\vh_v^{(r)} = \tau_r(l_v^{(r)})$ for $r = 0,1,\cdots,R$. In other words, the node embedding generated by the first $r$ branches of a PA-GNN model gathering information from its $r$-hop  neighborhood can be mapped to the labels generated at the $r$-th iteration of the 1-WL test. Let us assume that all the functions involved in generating the node embeddings, namely, $g^{(r)}(\cdot)$, $f_v^{(r)}(\cdot)$, and $\Theta_r(\cdot)$ are injective. 

Let us first verify the base case. For a single branch PA-GNN model with $R=0$, we have 
\begin{align*}
   \vh_v^{(0)} = \Theta_0\left(\vz_v^{(0)}\right) = \Theta_0\left(c_v^{(0)}(\vx_v)\right) = \Theta_0\left(c_v^{(0)}\left(l_v^{(0)}\right)\right) = \tau_0(l_v^{(0)})
\end{align*}
with $\tau_0(\cdot) = \Theta_0\circ c_v^{(0)}$. Next, for a two branch PA-GNN model with $R=1$, we have
\begin{align*}
    \vh_v^{(1)} = \Theta_1\left(\vz_v^{(0)},\vz_v^{(1)}\right) = \Theta_1\left(c_v^{(0)}\left(\vx_v\right),c_v^{(1)}\left(\left\{\vx_j,\; \forall j\in \mathcal{N}_v^{(1)}\right\}\right)\right).
\end{align*}
As the composition of injective functions is also injective, the above equation simplifies to 
\begin{align*}
    \vh_v^{(1)} = \rho_1\left(\vx_v,\left\{\vx_j,\; \forall j \in \mathcal{N}_v^{(1)}\right\}\right)
\end{align*}
for some injective function $\rho_1(\cdot)$. Since 
\[
l_v^{(1)} = \varphi\left(l_v^{(0)},\left\{l_j^{(0)}, \;\forall j \in \mathcal{N}_v^{(1)}\right\}\right),
\]
we have $\tau_1(\cdot) = \rho_1\circ \varphi^{-1}$.

Similarly, for a three branch PA-GNN model with $R=2$ we have
\begin{align*}
    \vh_v^{(2)} = \Theta_2\Bigl(c_v^{(0)}(\vx_v), c_v^{(1)}\left(\left\{\vx_j,\; \forall j\in \mathcal{N}_v^{(1)}\right\}\right),
    c_v^{(2)}\left(\left\{\vx_t,\;\forall t\in \mathcal{N}_j^{(1)}, \;\forall j \in \mathcal{N}_v^{(1)}\right\}\right)\Bigr).
\end{align*}
As $\Theta_r(\cdot)$ and $c_v^{(r)}(\cdot)$ are injective, $\vh_v^{(2)}$ can be expressed as
\begin{align}
    \vh_v^{(2)} = \rho_2\Bigl(\vx_v,\left\{\vx_j, \;\forall j\in \mathcal{N}_v^{(1)}\right\},
    \left\{\vx_t, \;\forall t\in \mathcal{N}_j^{(1)}, \forall \; j \in \mathcal{N}_v^{(1)}\right\}\Bigr)
    \label{eq:a1}
\end{align}
for some injective function $\rho_2(\cdot)$. In fact, the set $\left\{\vx_t, \;\forall t\in \mathcal{N}_j^{(1)}, \forall \; j \in \mathcal{N}_v^{(1)}\right\}$ in Equation~\eqref{eq:a1} represents embeddings at 2-hop neighbors of node $v$. 

In the 1st iteration of the 1-WL update, the labels depend on their 1-hop neighbors, and similarly, in the 2nd iteration, the labels depend on their 1-hop and 2-hop neighbors. The second iteration of the 1-WL update can be written as
\begin{equation}
\begin{aligned}
    l_v^{(2)} = \varphi\Bigl(\varphi\left(l_v^{(0)},\left\{l_u^{(0)}, \;\forall u \in \mathcal{N}_v^{(1)}\right\}\right),
    \left\{\varphi\left(l_j^{(0)},\left\{l_t^{(0)}, \;\forall t\in \mathcal{N}_j^{(1)}\right\}\right), \;\forall j\in \mathcal{N}_v^{(1)}\right\}\Bigr).
\end{aligned}
\end{equation}

As $\varphi(\cdot)$ in the 1-WL algorithm is an injective hashing function, we have
\begin{align*}
    l_v^{(2)} =  \varphi_2\Bigl(l_v^{(0)},\left\{l_j^{(0)}, \;\forall j\in \mathcal{N}_v^{(1)}\right\},
    \left\{l_t^{(0)}, \;\forall t\in \mathcal{N}_j^{(1)}, \;\forall j \in \mathcal{N}_v^{(1)}\right\}\Bigr)
\end{align*}
for some injective function $\varphi_2(\cdot)$. Hence, $\vh_v^{(2)} = \tau_2(l_v^{(2)})$ with $\tau_2(\cdot) = \rho_2\circ \varphi_2^{-1}$.

Next, we assume there exist an injective mapping function $\tau_{R-1}(\cdot)$ up to $R-1$ iterations such that $\vh_v^{(R-1)} = \tau_{R-1}(l_v^{(R-1)})$, and prove there exist an injective mapping for the $R$-th iteration. The embedding for node $v$ from a $R$ branch PA-GNN model is  
\begin{align*}
    \vh_v^{(R)} = \Theta_{R}\Bigl(c_v^{(0)}(\vx_v),\cdots,c_v^{(R-1)}\left(\left\{\vx_j, \;\forall j\in \mathcal{N}_v^{(R-1)}\right\}\right),
    c_v^{(R)}\left(\left\{\vx_j, \;\forall j\in \mathcal{N}_v^{(R)}\right\}\right)\Bigr),
\end{align*}
which can be alternatively represented as 
\begin{equation*}
    \vh_v^{(R)} =
    \Theta_{R}\left(\Theta_{R-1}^{-1}\circ \Theta_{R-1}\left(\mathcal{U}\right),c_v^{(R)}\left(\left\{\vx_j, \;\forall j\in \mathcal{N}_v^{(R)}\right\}\right)\right),
\end{equation*}
where 
\begin{equation*}
\mathcal{U} = \left\{ c_v^{(0)}(\vx_v),\cdots,c_v^{(R-1)}\left(\left\{\vx_j, \;\forall j\in \mathcal{N}_v^{(R-1)}\right\}\right)\right\}.
\end{equation*}
Therefore,
\begin{align*}
    \vh_v^{(R)} = \Theta_{R}\left(\Theta_{R-1}^{-1}\left(\vh_v^{(R-1)}\right),c_v^{(R)}\left(\left\{\vx_j, \;\forall j\in \mathcal{N}_v^{(R)}\right\}\right)\right).
\end{align*}
Replacing $\vh_v^{(R-1)}$ with $\tau_{R-1}(l_v^{(R-1)})$ and the node features $\vx_j$ with $l_j^{(0)}$, we get
\begin{align*}
    \vh_v^{(R)} = \Theta_{R}\Bigl(\Theta_R^{-1}\left(\tau_{R-1}\left (l_v^{(R-1)}\right)\right),
    c_v^{(R)}\left(\left\{l_j^{(0)}, \;\forall j\in \mathcal{N}_v^{(R)}\right\}\right)\Bigr).
\end{align*}
For some injective $\rho_R(\cdot)$, above equation simplifies to
\begin{align*}
    \vh_v^{(R)} = \rho_{R}\left(l_v^{(R-1)},\left\{l_j^{(0)}, \forall \; j\in \mathcal{N}_v^{(R)}\right\}\right).
\end{align*}
Similarly, the 1-WL update at the $R$-th iteration will be
\begin{align*}
    l_v^{(R)} = \varphi\left(l_v^{(R-1)},\left\{l_j^{(R-1)}, \;\forall j \in \mathcal{N}_v^{(1)}\right\}\right).
\end{align*}
We have seen above that at the $r$-th iteration, the 1-WL update depends on all the labels from nodes within its $r$-hop neighborhood. So, for some function $\varphi_{R-1}(\cdot)$, we can write $l_v^{(R)}$ as 
\begin{align*}
   l_v^{(R)} = \varphi_{R-1}\left(l_v^{(R-1)},\left\{l_j^{(0)},\;\forall j \in \mathcal{N}_v^{(R)}\right\}\right). 
\end{align*}
Hence, $\vh_v^{(R)} = \tau_R(l_v^{(R)})$ with $\tau_R(\cdot) = \rho_{R}\circ \varphi_{R-1}^{-1}$.

\subsection{Relating graph-level readouts from a PA-GNN model and the 1-WL algorithm} 

Next, we establish the second condition by analyzing the relation between the graph-level embeddings $\ve_1$ generated by a PA-GNN model and $\ve_2$ generated using the final labels (after the $R$-th iteration) from the 1-WL test for an entire graph. 

With PA-GNN models, we first obtain the embedding $\vs_G^{(r)}$ for an entire graph at each branch by the branch-level readout of the node representation vectors. Then the  embedding of an entire graph $\ve_1$ is computed through a global readout as
\begin{align*}
    &\ve_1 = \psi_1\left(\vs_G^{(0)},\vs_G^{(1)},\cdots,\vs_G^{(R)}\right) 
\end{align*}
with $\vs_G^{(r)} = \Omega\left(\left\{\vz_v^{(r)},\; \forall v \in \mathcal{V}\right\}\right)$. Here,
 $\Omega(\cdot)$ is the branch-level readout function that operates individually on each branch and $\psi_1(\cdot)$ is the global readout function that produces the graph embedding by pooling the branch-level embeddings. We assume $\Omega(\cdot)$ and $\psi_1(\cdot)$ to be injective. 

The graph representation $\ve_2$ generated from the node labels of the 1-WL algorithm after $R$ iterations is given by 
\begin{align*}
   \ve_2 = \psi_2\left(\left\{l_v^{(R)}, \; \forall v \in \mathcal{V}\right\}\right), 
\end{align*}
where $\psi_2(\cdot)$ is an injective graph pooling function.

We show that there always exist an injective function $\Psi(\cdot)$ that maps $\ve_1$ to $\ve_2$, i.e., $\ve_1 = \Psi(\ve_2)$.
To do so, let us express $\ve_1$ as
\begin{align*}
    \ve_1 = \psi_1\left(\left\{\Omega(\vz_v^{(0)}),\cdots,\Omega(\vz_v^{(R)}), \forall v \in \mathcal{V}\right\}\right).
\end{align*}
We introduce an injective function $\Delta = \psi_1 \circ \Omega$ such that
\begin{align*}
    \ve_1 = \Delta\left(\left\{\vz_v^{(0)},\cdots,\vz_v^{(R)}, \forall v \in \mathcal{V}\right\}\right)
    = \Delta\left(\Theta^{-1}\left(\left\{\vh_v^{(R)},\; \forall v \in \mathcal{V} \right\}\right)\right).
\end{align*}
As established previously, we have $\vh_v^{(R)} = \tau_R(l_v^{(R)})$. Thus
\begin{align*}
    \ve_1 = \Delta\left(\Theta^{-1}\left(\left\{\tau_R\left(l_v^{(R)}\right),\; \forall v \in \mathcal{V}\right\}\right)\right) = \Psi\left(\ve_2\right)
\end{align*}
with an injective function $\Psi(\cdot) = \Delta\circ\Theta_R^{-1}\circ\tau_R$.

In essence, all the functions involved in a PA-GNN model should be injective for it to be at least as powerful as the 1-WL test.

\section{Proof for Lemma 1} \label{app:lemma1}
For two distinct multisets $\mathfrak{X}_1 \neq \mathfrak{X}_2$, we next prove that the weighted summation of their linear mappings with a ReLU or Leaky-ReLU nonlinearity can be equal, i.e.,
\begin{align}
    \sum_{\vx \in \mathfrak{X}_1}\alpha_i \sigma(\mW\vx) = \sum_{\vx\in \mathfrak{X}_2}\beta_i \sigma(\mW\vx)
    \label{eq:relu_lem1}
\end{align}
for an arbitrary linear mapping $\mW$ and arbitrary weights $\alpha_i$ for $i = 1,2,\cdots,|\mathfrak{X}_1|$ and $\beta_i$ for $i = 1,2,\cdots,|\mathfrak{X}_2|$. Here the nonlinearity $\sigma(\cdot)$ is either $\texttt{ReLU}(x)$ or $\texttt{L-ReLU}(x)$, where ReLU is defined as 
\[\texttt{ReLU}(x) =
    \left\{
	\begin{array}{ll}
		x  & \mbox{if } x > 0 \\
		0 & \mbox{if } x \leq 0
	\end{array}
\right.
\]
and Leaky-ReLU is defined as 
\[
\texttt{L-ReLU}(x) =
    \left\{
	\begin{array}{ll}
		x  & \mbox{if } x > 0 \\
	cx & \mbox{if } x \leq 0 \textrm{ for } c>0.
	\end{array}
\right.
\]

\subsection{The case with ReLU}
By definition, $\texttt{ReLU}(\mW\vx)$ is elementwise positive or zero depending on $\mW\vx$ being elementwise positive or negative, respectively. Let us introduce the symbols $\succ$ and $\prec$ to, respectively, denote elementwise greater than or elementwise less than inequalities for vectors.

When $\mW\vx \prec 0$, $\texttt{ReLU}(\mW\vx) = 0$ for all $\vx \in \mathfrak{X}_1 \cup \mathfrak{X}_2$. Then
\begin{align*}
    \sum_{\vx\in \mathfrak{X}_1}\alpha_i 
\texttt{ReLU}(\mW\vx) = \sum_{\vx\in \mathfrak{X}_2}\beta_i 
\texttt{ReLU}(\mW\vx) = 0
\end{align*}
irrespective of the weights $\alpha_i$ and $\beta_i$. When $\mW\vx\succ 0$, $\texttt{ReLU}(\mW\vx) = \mW\vx$. Thus
\begin{align*}
   \sum_{\vx\in \mathfrak{X}_1}\alpha_i \texttt{ReLU}(\mW\vx) = \mW\left(\sum_{\vx\in \mathfrak{X}_1}\alpha_i \vx\right) \textrm{ and }
   \sum_{\vx\in \mathfrak{X}_2}\beta_i \texttt{ReLU}(\mW\vx) = \mW\left(\sum_{\vx\in \mathfrak{X}_2}\beta_i \vx\right). 
\end{align*}
Hence, we can see that if the weighted sums $\sum_{\vx\in \mathfrak{X}_1}\alpha_i\vx = \sum_{\vx\in \mathfrak{X}_2}\beta_i\vx$, their linear mappings followed by the ReLU nonlinearity can be equal.

\subsection{The case with Leaky-ReLU}
For $\mW\vx\succ 0$, the proof remains exactly the same as with the case of ReLU as ReLU and Leaky-ReLu are the same for positive inputs. When $\mW\vx\prec 0$, $\texttt{L-ReLU}(\mW\vx) = c\mW\vx$. Since scalar multiplication does not affect linearity, the above argument holds.

\textbf{Example.} Consider $\mathfrak{X}_1 = \{2,1,4\}$ with weights $\alpha = \{1,-1,0.25\}$ and $\mathfrak{X}_2 = \{6,4\}$ with $\beta = \{1,-1\}$. The multisets are different, but their weighted sums are equal. For any linear transformation $\mW$, their weighted sum followed by ReLU or Leaky-ReLU are the same. 

\section{Proof for Lemma 2} \label{app:lemma2}

From the conditions provided in Theorem~1, we choose injective functions for the aggregation function $f_v^{(r)}(\cdot)$ and the transformation function $g^{(r)}(\cdot)$. Therefore, the intermediate embeddings $\vz_v^{(r)}$ produced for two different feature vectors are distinct. 

Consider two distinct intermediate embeddings $\vz_1$ and $\vz_2$ generated from the same branch for two non-isomorphic graphs. Let us denote the node representation vectors with attention as $\tilde{\vz}_i = \alpha_i\vz_i$, where $\alpha_i = \texttt{ATTENTION}(\vw,\vz_i)$ with an arbitrary vector $\vw$. The attention mechanism is defined as
\begin{align*}
    \alpha_i &= \texttt{ATTENTION}(\vw, \vz_i) = \frac{1}{n_i}\exp(\beta_i),
\end{align*}
where  $\beta_i = \texttt{ReLU}\left(\langle\vw, \vz_i\rangle\right)$. Here, we have introduced the normalization factor $n_i$ due to the softmax operation.

We show that for distinct non-zero vectors $\vz_1 \neq \vz_2$ (element-wise inequality), the attention mechanism preserves injectivity, i.e., $\tilde{\vz}_1 \neq \tilde{\vz}_2$. In particular, we show that $\tilde{\vz}_1 \neq \tilde{\vz}_2$ for the two cases with vectors $\vz_1$ and $\vz_2$ being linearly independent and dependent but not identical.

To begin with, we analyze the conditions under which $\tilde{\vz}_1 = \tilde{\vz}_2$. 
Assuming that $\tilde{\vz}_1 = \tilde{\vz}_2$, we have 
\begin{align*}
    \frac{1}{n_1}\exp(\beta_1)\vz_1 = \frac{1}{n_2}\exp(\beta_2)\vz_2, 
\end{align*}
which implies that $\vz_1 = p\vz_2$ with $p = n_1 \exp(\beta_2 - \beta_1)/n_2$. In other words, for $\tilde{\vz}_1 = \tilde{\vz}_2$, the vectors $\vz_1$ and $\vz_2$ have to be linear dependent. 

Now, we inspect the conditions on $p$ under which $\vz_1$ and $\vz_2$ are linearly dependent, i.e., $\vz_2 = p\vz_1$ that lead to $\tilde{\vz}_1 = \tilde{\vz}_2$. 
When $\vz_2 = p\vz_1$, we have
\begin{align*}
   \exp(\beta_1)\vz_1 = p\exp(\beta_2)\vz_1.
\end{align*}
Since the scaling factors $n_1$ and $n_2$ are positive, with a slight abuse of notation, we have absorbed them in the constant $p$. 
Substituting $\beta_i$, we get
\begin{align*}
    \exp(\texttt{ReLU}(\langle\vw,\vz_1\rangle))\vz_1 = p\exp(\texttt{ReLU}(\langle\vw,p\vz_1\rangle))\vz_1.
\end{align*}

When $p>0$, using the fact that the scaled ReLU can be written as $\texttt{ReLU}(px) = p\texttt{ReLU}(x)$, we obtain the equation
\begin{align*}
    b - pb^p = 0
\end{align*}
with $b = \exp(\texttt{ReLU}(\langle\vw, \vz_1\rangle)).$ The general solution to above equation is given by $p = \frac{J_n(b\log(b))}{\log(b)}$, where $J_n(\cdot)$ is the analytic continuation of the product log function~\cite{Wn}. However, since $b>1$ as $b = \exp(\texttt{ReLU}(\langle\vw, \vz_1\rangle))$, the only real solution to the equation is $p=1$. This, in other words, means that all the entries of the vectors $\vz_1$ and $\vz_2$ are identical.

When $p<0$, we have the following two cases:
\begin{itemize}
    \item When $p<0$ and $\langle\vw,\vz_1\rangle<0$, the equation $\tilde{\vz}_1 = \tilde{\vz}_2$ becomes
\begin{align*}
    \vz_1 = p\exp(p\langle\vw,\vz_1\rangle)\vz_1,
\end{align*}
which can be equivalently written as
\begin{align*}
\vz_1(p\exp(p\langle\vw,\vz_1\rangle)-1) = 0.
\end{align*}
Since $p <0$, we have $\exp(p\langle\vw,\vz_1\rangle) \geq 0$. Therefore, $p\exp(p\langle\vw,\vz_1\rangle) \neq 1$.
 
\item When $p<0$ and $\langle\vw,\vz_1\rangle>0$, we can write the equation $\tilde{\vz}_1 = \tilde{\vz}_2$ as
\begin{align*}
    \vz_1\exp(\langle\vw,\vz_1\rangle) = p\vz_1, 
\end{align*}
or, equivalently as
\begin{align*}
\exp(\langle\vw,\vz_1\rangle) = p,
\end{align*}
which is not true as $\exp(\langle\vw,\vz_1\rangle) \geq 0$ and $p<0$.  
\end{itemize}
Hence, a $p<0$ for which $\tilde{\vz}_1 = \tilde{\vz}_2$ does not exist.
In summary, the only possible condition for $\tilde{\vz}_1 = \tilde{\vz}_2$ is when $\vz_1 = p\vz_2$ with $p=1$. Thus, the attention mechanism preserves injectivity.

\section{Proof for Lemma 3} \label{app:lemma3}

We show that the mean or max readout functions reduces the discriminative power of a PA-GNN model when the graphs under test are regular and have the same node type.

Consider a graph $\mathcal{G} = (\mathcal{V,E})$ with $N$ nodes. Recall that a PA-GNN model produces the following node and graph embeddings at the $r$-th branch:
\begin{align*}
    \vz_v^{(r)} = g^{(r)}\left(f_v^{(r)}\left(\mathfrak{X}_v^{(r)}\right)\right) \text{ and  } \, \vs_G^{(r)} = \Omega\left(\left\{\vz_v^{(r)}, \forall v \in \mathcal{V}\right\}\right),
\end{align*}
where $\vz_v^{(r)}$ is the intermediate embedding generated at the $r$-th branch for node $v \in \mathcal{V}$ and $\vs_G^{(r)}$ represents the branch-level representation for the graph at the $r$-th branch. Let $\mathfrak{X}_v^{(r)} = \left\{\vx_j, \; \forall \; j \in \mathcal{N}_v^{(r)}\right\}$ be a multiset of features of the $r$-hop neighboring nodes of node $v$. We denote $f_v^{(r)}(\mathfrak{X}_v^{(r)}) = \vb_v^{(r)}$ with $\mB^{(r)} = [\vb_1^{(r)},\cdots,\vb_N^{(r)}]$. Alternatively, we represent the intermediate embeddings as 
\begin{align*}
    \mZ^{(r)} = g^{(r)}\left(\mB^{(r)}\right),
\end{align*}
where $\mZ^{(r)} = [\vz_1^{(r)},\cdots,\vz_N^{(r)}]\rT$.

Consider the graphs $\mathcal{G}_1 = (\mathcal{V}_1,\mathcal{E}_1)$ and $\mathcal{G}_2 = (\mathcal{V}_2,\mathcal{E}_2)$ with nodes $|\mathcal{V}_1| = N_1$, $|\mathcal{V}_2| = N_2$, respectively, with $N_1 \neq N_2$. Let $\mX_1 \in \mathbb{R}^{N_1\times d}$ and $\mX_2\in \mathbb{R}^{N_2\times d}$ be the node attributes associated with $\mathcal{G}_1$ and $\mathcal{G}_2$, respectively. 
Suppose the graphs $\mathcal{G}_1$ and $\mathcal{G}_2$ are regular and have the same node type, then the neighborhood multisets are the same for all the nodes, i.e., $\mathfrak{X}_v^{(r)} = \mathfrak{X}_u^{(r)}, \forall v \in \mathcal{V}_1 \textrm{ and } u \in \mathcal{V}_2$. This implies that the feature matrix $\mB_1^{(r)}$ for graph $\mathcal{G}_1$ and $\mB_2^{(r)}$ for graph $\mathcal{G}_2$ generated after the neighborhood aggregation $f^{(r)}(\cdot)$ at the $r$-th branch have the same rows, i.e., $\vb_v^{(r)} = \vb_u^{(r)} = \vb, \; \forall v\in \mathcal{V}_1, u \in \mathcal{V}_2.$ Thus, $\mB_i^{(r)} = {\bf 1}_{N_i}\vb\rT$, where $\textbf{1}_{N_i} \in \mathbb{R}^{N_i}$ is an all-one vector of length $N_i$. Then the intermediate embeddings for graph $\mathcal{G}_i$ are given by
\begin{align*}
    \mZ_i^{(r)} = g^{(r)}\left(\mB_i^{(r)}\right) = g^{(r)}\left({\bf 1}_{N_i}\vb\rT\right). 
\end{align*}
As the transformation function $g^{(r)}(\cdot)$ is injective (as prescribed by Theorem 1) and designed using MLPs, it will produce the same output for the same input vector. Consequently, we can represent it as
\begin{align*}
    \mZ_i^{(r)} = g^{(r)}\left(\mB_i^{(r)}\right) = g^{(r)}\left({\bf 1}_{N_i}\vb\rT\right)
    = {\bf 1}_{N_i}g^{(r)}\left(\vb\rT\right) = {\bf 1}_{N_i}\vy\rT,
\end{align*}
where $\vy = g^{(r)}(\vb)$. We can see that all the rows in the intermediate representation matrix $\mZ_i^{(r)}$ are $\vy$, and are independent of the graph index $i$. 

Next, the branch-level graph pooling $\Omega(\cdot)$ operates on these intermediate embeddings. For the graph $\mathcal{G}_i$, the branch-level readout is
\begin{align*}
   \vs_{G_i}^{(r)} = \Omega\left(\left\{\vz_v^{(r)}, \forall v \in \mathcal{V}_i\right\}\right)
   = \Omega\left(\{\vy_1,\vy_2,\cdots,\vy_{N_i}\}\right) = \Omega\left(\{\vy,\vy,\cdots,\vy\}\right), 
\end{align*}
where all the vectors $\vy_i = \vy$. It is now easy to observe that, if $\Omega(\cdot)$ = $\texttt{mean}(\cdot)$ or $\texttt{max}(\cdot)$, the branch-level readout would produce the same output
\begin{align*}
  \vs_{G_i}^{(r)} = \texttt{mean}\left(\{\vy_1,\vy_2,\cdots,\vy_{N_i}\}\right)
  = \texttt{max}(\{\vy_1,\vy_2,\cdots,\vy_{N_i}\}) = \vy.
\end{align*}
Furthermore, as all the branch-level readouts for both the graphs are the same, we cannot differentiate the graphs under test from the global readout as the graph embeddings are identical
$$\ve_{G_i} = \psi\left(\vs_{G_i}^{(0)},\cdots,\vs_{G_i}^{(R)}\right) = \psi(\vy,\vy,\cdots,\vy),$$
where the final graph representation is also independent of the graph index $i$.

\noindent\textbf{Example.} Suppose we use integer powers of the normalized adjacency matrix, i.e., $\tilde{\mA}_i = \mD_i^{-1/2}\mA_i\mD_i^{-1/2}$ with $\mD_i$ being the diagonal degree matrix of~$\mathcal{G}_i$, as a graph operator for neighborhood aggregations. Then we have $\mB_i^{(r)} = \tilde{\mA}_i^r\mX_i$ at the $r$-th branch, where $\tilde{\mA}_i$ is the normalized adjacency matrix of graph $\mathcal{G}_i$ and the feature matrix $\mX_i = [\vx\;\;\vx\;\;\cdots\;\;\vx]\rT = {\bf 1}_{N_i}\vx\rT$.

For a regular graph, with $\mD_i = t\mI$, we have $\tilde{\mA}_i = t^{-1}\mA_i$. Here, $\mI$ is the identity matrix. The adjacency matrix $\mA_i$ corresponding to a regular graph $\mathcal{G}_i$ has an eigenvalue $t$ corresponding to its all-one eigenvector, i.e., $\mA_{i}{\bf 1}_{N_i} = t{\bf 1}_{N_i}$. Thus we have $\tilde{\mA}_i{\bf 1}_{N_i} = tt^{-1}{\bf 1}_{N_i} = {\bf 1}_{N_i}$ due to which the expression for $\mB_i^{(r)}$ simplifies to
\begin{align*}
   \mB_i^{(r)} = \tilde{\mA}_i\mX_i = \tilde{\mA}_i{\bf 1}_{N_i}\vx\rT = {\bf 1}_{N_i}\vx\rT.
\end{align*}
Hence
\begin{align*}
    \mZ_i^{(r)} = g^{(r)}\left(\mB_i^{(r)}\right) = g^{(r)}\left({\bf 1}_{N_i}\vx\rT\right)
    = {\bf 1}_{N_i}g^{(r)}\left(\vx\rT\right) = {\bf 1}_{N_i}\vy\rT
\end{align*}
with all the intermediate embeddings stacked in $\mZ_i^{(r)}$ are the same vector $\vy$, which is independent of the graph index $i$.

\begin{table*}[ht]
    \centering
    \caption{{Dataset details.}}
    \label{Tab:stats}
    \renewcommand{\arraystretch}{1.2}
    \scalebox{1}{
    \begin{tabular}{*6c}
        \hline
        {}   & Graphs & Classes & Avg. nodes & Avg. edges & Input feat dim\\
        \hline
        D\&D & 1178 & 2 & 284.32 & 715.66 & 89\\
        NCI1 & 4110 & 2 & 29.87 & 32.3 & 37\\
        PROTEINS & 1113 & 2 & 39.06 & 72.82 & 3\\
        ENZYMES & 600 & 6 & 32.63 & 62.14 & 21\\
        OGBG-MOLHIV & 41127 & 2& 25.5 & 124.2 & 9\\
        \hline
        IMDB-BINARY & 1000 & 2 & 19.77 & 96.53 & 1\\
        IMDB-MULTI & 1500 & 3 & 13.00  & 65.94 & 1\\
        REDDIT-BINARY & 2000 & 2 & 429.63 & 497.75 & 1\\
        REDDIT-MULTI & 5000 & 5 & 508.52 & 594.87 & 1\\
        COLLAB & 5000 & 3 & 74.49 & 2457.78 & 1\\
        \hline
        OHSU & 79 & 2 &82.01 &199.66 & 1\\
        PEKING-1 & 85 & 2 & 39.31 &77.35 &1\\
        \hline
    \end{tabular}}
\end{table*}

\begin{table*}[ht]
    \centering
    \caption{{Hyperparameters the yield the best performance of SPIN. $|T|$ indicates the complete training set.}}
    \label{Tab:hyper}
    \renewcommand{\arraystretch}{1.2}
    \scalebox{0.89}{
    \begin{tabular}{c  *5c}
        \hline
        {}   & Batch size & \vtop{\hbox{\strut Number of}\hbox{\strut branches ($R$)}} & Intermediate dimension & Learning rate & L2 regularization\\
        \hline
        DD & \{16,32\} & \{2,3,4\} & \{16,32,64\} & 5$\times10^{-3}$ & No\\
        NCI1 & \{64,128\} & \{1,2\} & \{16,32\} & 1$\times10^{-3}$ & No\\
        PROTEINS & \{16,32,64\} & \{2,3\} & \{8,16\} & 1$\times10^{-3}$ & No\\
        ENZYMES & \{8,16,32\} & \{2,3\} & \{8,16\} & 1$\times10^{-3}$ & 1$\times10^{-3}$\\
        OGBG-OLHIV & \{32,64,128\} & \{2,3\} & \{16,32,64,128\} & 1$\times10^{-4}$ & No\\
        \hline
        IMDB-BINARY & \{16,32\} & \{2,3,4\} & \{8,16,32\} & 5$\times10^{-3}$ & No\\
        IMDB-MULTI & \{16,32,64\} & \{2,3,4\} & \{8,16,32\} & 5$\times10^{-3}$ & No\\
        REDDIT-BINARY & \{32,64,128\} & \{3,4\} & \{8,16\} & 5$\times10^{-3}$ & No\\
        REDDIT-MULTI & \{64,128\} & \{3,4\} & \{8,16\} & 5$\times10^{-3}$ & No\\
        COLLAB & \{32,64,128\} & \{2,3,4\} & \{8,16,32,64\} & 5$\times10^{-3}$ & 5$\times10^{-4}$\\
        \hline
        OHSU & \{8,16,$|T|$\} & \{8,16\} & \{8,16,32\} & 5$\times10^{-3}$ & 5$\times10^{-4}$\\
        PEKING-1 & \{8,16,$|T|$\} & \{4,8,16\} & \{8,16\} & 5$\times10^{-3}$ & 5$\times10^{-4}$\\
        \hline
    \end{tabular}}
\end{table*}

\newpage
\bibliographystyle{unsrt}
\bibliography{sample}

\begin{thebibliography}{10}

\bibitem{GDL}
{Bronstein MM, Bruna J, LeCun Y, Szlam A, Vandergheynst P.}
\newblock {Geometric deep learning: going beyond euclidean data}.
\newblock {\em IEEE Signal Processing Magazine}, {34}({4}):{18--42}, {July}
  {2017}.

\bibitem{gamma20spm}
{Gama F, Isufi E, Leus G, Ribeiro A.}
\newblock {Graphs, convolutions, and neural networks: From graph filters to
  graph neural networks}.
\newblock {\em {IEEE Signal Processing Magazine.}}, {37}({6}):{128--138},
  {Sept.} {2020}.

\bibitem{decagon}
{Zitnik M, Agrawal M, Leskovec J.}
\newblock {Modeling polypharmacy side effects with graph convolutional
  networks}.
\newblock {\em {Bioinformatics}}, {34}({13}):{i457--i466}, {July} {2018}.

\bibitem{MoIGAN}
{De Cao N, Kipf T.}
\newblock {An implicit generative model for small molecular graphs}.
\newblock {\em {arXiv preprint arXiv:1805.11973}}, {May} {2018}.

\bibitem{phy}
{Sanchez-Gonzalez A, Godwin J, Pfaff T, Ying R, Leskovec J, Battaglia P.}
\newblock {Learning to simulate complex physics with graph networks}.
\newblock In {\em Proceedings of the International Conference on Machine
  Learning}, {Vienna, Austria}, {July} {2020}.

\bibitem{finance1}
{Matsunaga D, Suzumura T, Takahashi T.}
\newblock {Exploring graph neural networks for stock market predictions with
  rolling window analysis}.
\newblock {\em {arXiv preprint arXiv:1909.10660}}, {Sept.} {2019}.

\bibitem{finance2}
{Weber M, Domeniconi G, Chen J, Weidele DK, Bellei C, Robinson T, Leiserson
  CE.}
\newblock {Experimenting with graph convolutional networks for financial
  forensics}.
\newblock {\em {arXiv preprint arXiv:1908.02591}}, {July} {2019}.

\bibitem{wicom}
{He S, Xiong S, Ou Y, Zhang J, Wang J, Huang Y, Zhang Y.}
\newblock {An Overview on the Application of Graph Neural Networks in Wireless
  Networks}.
\newblock {\em {arXiv preprint arXiv:2107.03029}}, {July} {2021}.

\bibitem{opt}
{Dai H, Khalil EB, Zhang Y, Dilkina B, Song L.}
\newblock {Learning combinatorial optimization algorithms over graphs}.
\newblock {\em {arXiv preprint arXiv:1704.01665}}, {Apr.} {2017}.

\bibitem{dong_gsp}
{Dong X, Thanou D, Toni L, Bronstein M, Frossard P.}
\newblock {Graph signal processing for machine learning: A review and new
  perspectives}.
\newblock {\em IEEE Signal Processing Magazine}, {37}({6}):{117--127}, {Oct.}
  {2020}.

\bibitem{graphBook}
{Hamilton WL.}
\newblock {Graph representation learning.}
\newblock {\em {Synthesis Lectures on Artifical Intelligence and Machine
  Learning.}}, {14}({3}):{1--59}, {Sept.} {2020}.

\bibitem{GCN}
{Kipf TN, Welling M.}
\newblock {Semi-supervised classification with graph convolutional networks}.
\newblock In {\em Proceedings of the International Conference on Learning
  Representations}, {Toulon, France}, {Apr.} {2017}.

\bibitem{GraphSAGE}
{Hamilton WL, Ying R, Leskovec J.}
\newblock {Inductive representation learning on large graphs}.
\newblock In {\em Advances in Neural Information Processing Systems},
  {California, United States}, {Dec.} {2017}.

\bibitem{GAT}
{Veličković P, Cucurull G, Casanova A, Romero A, Lio P, Bengio Y.}
\newblock {Graph attention networks}.
\newblock In {\em Proceedings of the International Conference on Learning
  Representations}, {Vancouver, Canada}, {Apr.} {2018}.

\bibitem{GIN}
{Xu K, Hu W, Leskovec J, Jegelka S.}
\newblock {How powerful are graph neural networks?}
\newblock In {\em Proceedings of the International Conference on Learning
  Representations}, {New Orleans, United States}, {May} {2019}.

\bibitem{Li18deepGCN}
{Li Q, Han Z, Wu XM.}
\newblock {Deeper insights into graph convolutional networks for
  semi-supervised learning}.
\newblock In {\em Proceedings of the AAAI Conference on Artificial
  Intelligence}, {New Orleans, United States}, {Apr.} {2018}.

\bibitem{SAG}
{Lee J, Lee I, Kang J.}
\newblock {Self-attention graph pooling}.
\newblock In {\em Proceedings of the International Conference on Machine
  Learning}, {California, United States}, {June} {2019}.

\bibitem{diffpool}
{Ying R, You J, Morris C, Ren X, Hamilton WL, Leskovec J.}
\newblock { Hierarchical graph representation learning with differentiable
  pooling}.
\newblock In {\em Advances in Neural Information Processing Systems},
  {Montreal, Canada}, {Dec.} {2018}.

\bibitem{DGCNN}
{Zhang, M., Cui, Z., Neumann, M. and Chen, Y.}
\newblock {An end-to-end deep learning architecture for graph classification}.
\newblock In {\em Proceedings of the AAAI Conference on Artificial
  Intelligence}, {New Orleans, United States}, {Feb.} {2018}.

\bibitem{ECC}
{Simonovsky M, Komodakis N}.
\newblock {Dynamic Edge-Conditioned Filters in Convolutional Neural Networks on
  Graphs}.
\newblock In {\em Proceedings of the IEEE conference on Computer Vision and
  Pattern Recognition (CVPR)}, {Hawaii, United States}, {July} {2017}.

\bibitem{SIGN}
{Rossi E, Frasca F, Chamberlain B, Eynard D, Bronstein M, Monti F.}
\newblock {SIGN: Scalable Inception Graph Neural Networks}.
\newblock {\em {arXiv preprint arXiv:2004.11198}}, {Apr.} {2020}.

\bibitem{GAMLP}
{Chen L, Chen Z, Bruna J.}
\newblock {On graph neural networks versus graph-augmented {MLPs}}.
\newblock In {\em Proceedings of the International Conference on Learning
  Representations}, {Vienna, Austria}, {Mar.} {2021}.

\bibitem{inception}
{Szegedy C, Vanhoucke V, Ioffe S, Shlens J, Wojna Z.}
\newblock {Rethinking the inception architecture for computer vision}.
\newblock In {\em Proceedings of the IEEE conference on Computer Vision and
  Pattern Recognition (CVPR)}, {Nevada, United States}, {June} {2016}.

\bibitem{Tremblay17graphfilterbank}
{Tremblay N, Gonçalves P, Borgnat P.}
\newblock {Design of graph filters and filterbanks}.
\newblock In {\em {Cooperative and Graph Signal Processing}}, pages {299--324}.
  {Elsevier}, {2018}.

\bibitem{Morris}
{Morris C, Ritzert M, Fey M, Hamilton WL, Lenssen JE, Rattan G, Grohe M.}
\newblock {Weisfeiler and leman go neural: Higher-order graph neural networks}.
\newblock In {\em Proceedings of the AAAI Conference on Artificial
  Intelligence}, {Hawaii, United States}, {Jan.} {2019}.

\bibitem{WL_1968}
{Leman AA, Weisfeiler B.}
\newblock {A reduction of a graph to a canonical form and an algebra arising
  during this reduction}.
\newblock {\em {Nauchno-Technicheskaya Informatsia}}, {2}({9}):{12--16},
  {1968}.

\bibitem{babai}
{Babai L, Kucera L.}
\newblock {Canonical labelling of graphs in linear average time}.
\newblock {\em {20th Annual Symposium on Foundations of Computer Science}},
  pages {39--46}, {1979}.

\bibitem{kernel1}
{Shervashidze N, Schweitzer P, Van Leeuwen EJ, Mehlhorn K, Borgwardt KM.}
\newblock {{Weisfeiler-Lehman} graph kernels}.
\newblock {\em {Journal of Machine Learning Research.}}, {12}({9}):{}, {Sept.}
  {2011}.

\bibitem{social}
{Yanardag P, Vishwanathan SV. }.
\newblock {Deep graph kernels}.
\newblock In {\em {Proceedings of the 21th ACM SIGKDD international conference
  on knowledge discovery and data mining}}, pages {1365--1374}, {Aug.} {2015}.

\bibitem{JK}
{Xu K, Li C, Tian Y, Sonobe T, Kawarabayashi KI, Jegelka S. }.
\newblock {Representation learning on graphs with jumping knowledge networks}.
\newblock In {\em Proceedings of the International Conference on Machine
  Learning}, {Stockholm, Sweden}, {July} {2018}.

\bibitem{garey}
{Garey MR.}
\newblock {A Guide to the Theory of NP-Completeness}.
\newblock {\em {Computers and intractability.}}, {1979}.

\bibitem{proteins}
{Borgwardt KM, Ong CS, Schönauer S, Vishwanathan SV, Smola AJ, Kriegel HP. }.
\newblock {Protein function prediction via graph kernels}.
\newblock {\em {Bioinformatics}}, {21}({suppl\_1}):{i47--i56}, {June} {2005}.

\bibitem{survey}
{Wu Z, Pan S, Chen F, Long G, Zhang C, Philip SY. }.
\newblock {A comprehensive survey on graph neural networks}.
\newblock {\em {IEEE transactions on neural networks and learning systems.}},
  {32}({1}):{4--24}, {Jan.} {2021}.

\bibitem{DD}
{Dobson PD, Doig AJ. }.
\newblock {Distinguishing enzyme structures from non-enzymes without
  alignments}.
\newblock {\em {Journal of Molecular Biology}}, {330}({4}):{771--783}, {July}
  {2003}.

\bibitem{nci1}
{Wale N, Watson IA, Karypis G. }.
\newblock {Comparison of descriptor spaces for chemical compound retrieval and
  classification}.
\newblock {\em {Knowledge and Information Systems}}, {14}({3}):{347--375},
  {Mar.} {2008}.

\bibitem{enzymes}
{Schomburg I, Chang A, Ebeling C, Gremse M, Heldt C, Huhn G, Schomburg D. }.
\newblock {BRENDA, the enzyme database: updates and major new developments}.
\newblock {\em {Nucleic Acids Research}}, {32}({suppl\_1}):{D431--D433}, {Jan.}
  {2004}.

\bibitem{ogb}
{Hu W, Fey M, Zitnik M, Dong Y, Ren H, Liu B, Catasta M, Leskovec J.}
\newblock {Open graph benchmark: Datasets for machine learning on graphs}.
\newblock {\em {arXiv preprint arXiv:2005.00687}}, {May} {2020}.

\bibitem{brain}
{Pan S, Wu J, Zhu X, Long G, Zhang C.}
\newblock {Task sensitive feature exploration and learning for multitask graph
  classification}.
\newblock {\em {IEEE transactions on cybernetics.}}, {47}({3}):{744--758},
  {Mar.} {2016}.

\bibitem{public}
{Kersting K, Kriege NM, Morris C, Mutzel P, Neumann M.}
\newblock {Benchmark Data Sets for Graph Kernels}, {2016}.

\bibitem{Fair}
{Errica F, Podda M, Bacciu D, Micheli A.}
\newblock {A fair comparison of graph neural networks for graph
  classification}.
\newblock In {\em Proceedings of the International Conference on Learning
  Representations}, {Addis Ababa, Ethiopia}, {Apr.} {2020}.

\bibitem{moleculenet}
{Wu Z, Ramsundar B, Feinberg EN, Gomes J, Geniesse C, Pappu AS, Leswing K,
  Pande V.}
\newblock {MoleculeNet: a benchmark for molecular machine learning}.
\newblock {\em {Chemical science.}}, {9}({2}):{513--530}, {Mar.} {2018}.

\bibitem{MFT1}
{Ralaivola L, Swamidass SJ, Saigo H, Baldi P.}
\newblock {Graph kernels for chemical informatics}.
\newblock {\em {Neural Networks}}, {18}({8}):{1093--1110}, {Sept.} {2005}.

\bibitem{MFT2}
{Luzhnica E, Day B, Liò P. }.
\newblock {On graph classification networks, datasets and baselines}.
\newblock In {\em Proceedings of the International Conference on Machine
  Learning}, {California, United States}, {June} {2019}.

\bibitem{deepsets}
{Zaheer M, Kottur S, Ravanbakhsh S, Poczos B, Salakhutdinov R, Smola A.}
\newblock {Deep sets}.
\newblock In {\em Advances in Neural Information Processing Systems},
  {California, United States}, {Dec.} {2017}.

\bibitem{Wn}
{Corless RM, Gonnet GH, Hare DE, Jeffrey DJ, Knuth DE.}
\newblock {On the Lambert W function}.
\newblock {\em {Advances in Computational mathematics.}}, {5}({1}):{329--359},
  {Dec.} {1996}.

\end{thebibliography}

\end{document}